\documentclass[runningheads]{llncs}
\usepackage{eccv}
\usepackage{eccvabbrv}
\usepackage{graphicx}
\usepackage{booktabs}
\usepackage[accsupp]{axessibility}
\usepackage[pagebackref,breaklinks,colorlinks,citecolor=eccvblue]{hyperref}
\usepackage{orcidlink}
\newcommand{\red}[1]{{\color{red}#1}}

\newcommand{\Review}[1]{}
\newcommand{\alex}[1]{}

\usepackage{soul}
\setuldepth{foobar}

\newcommand{\R}{\mathbb{R}}
\usepackage{mathtools}
\usepackage{multirow}
\usepackage{rotating}
\usepackage{array}
\usepackage{amssymb}
\usepackage[table]{xcolor}
\definecolor{tabred}{rgb}{1, 0.7, 0.7}
\definecolor{taborange}{rgb}{1, 0.85, 0.7}
\definecolor{tabyellow}{rgb}{1, 1, 0.7}
\definecolor{tabgray}{rgb}{0.75, 0.75, 0.75}
\renewcommand{\red}[1]{\cellcolor{tabred}#1}
\newcommand{\orange}[1]{\cellcolor{taborange}#1}
\newcommand{\yellow}[1]{\cellcolor{tabyellow}#1}

\newcommand{\highlightinline}[2]{\begingroup\setlength{\fboxsep}{1pt}\colorbox{#1}{#2}\endgroup}
\newcommand{\paragraphtitle}[1]{\noindent \textbf{#1}}

\usepackage{stfloats}

\begin{document}
\title{Fisheye3R: Adapting Unified 3D Feed-Forward Foundation Models to Fisheye Lenses}
\titlerunning{Fisheye3R}
\author{
    Ruxiao Duan\inst{1,2} \and
    Erin Hong\inst{2} \and
    Dongxu Zhao\inst{2} \and
    \\
    Eric Turner\inst{2} \and
    Alex Wong\inst{1} \and
    Yunwen Zhou\inst{2}
}
\authorrunning{R.~Duan et al.}
\institute{
    Yale University \\
    \email{\{ruxiao.duan, alex.wong\}@yale.edu}
    \and
    Google \\
    \email{\{ruxiao, honge, dongxuz, elturner, verse\}@google.com}
}

\maketitle

\begin{abstract}

Feed-forward foundation models for multi-view 3-dimensional (3D) reconstruction have been trained on large-scale datasets of perspective images; when tested on wide field-of-view images, e.g., from a fisheye camera, their performance degrades. This degradation arises from changes in spatial arrangements of pixels induced by the non-linear projection model that maps 3D points onto the 2D image plane. While one may surmise that training on fisheye images would resolve this problem, there are far fewer fisheye images with ground truth than perspective images, which limits generalization. To enable inference on imagery exhibiting high radial distortion, we propose \textit{Fisheye3R}, a novel adaptation framework that extends these multi-view 3D reconstruction foundation models to natively accommodate fisheye inputs without performance regression on perspective images. To address the scarcity of fisheye images and ground truth, we introduce flexible learning schemes that support self-supervised adaptation using only unlabeled perspective images and supervised adaptation without any fisheye training data. Extensive experiments across three foundation models, including VGGT, $\pi^3$, and MapAnything, demonstrate that our approach consistently improves camera pose, depth, point map, and field-of-view estimation on fisheye images. Code is available at https://github.com/android-xr/fisheye3r.

\end{abstract}
\section{Introduction}
\label{sec:introduction}

\begin{figure*}[t]
    \centering
    \includegraphics[width=\linewidth]{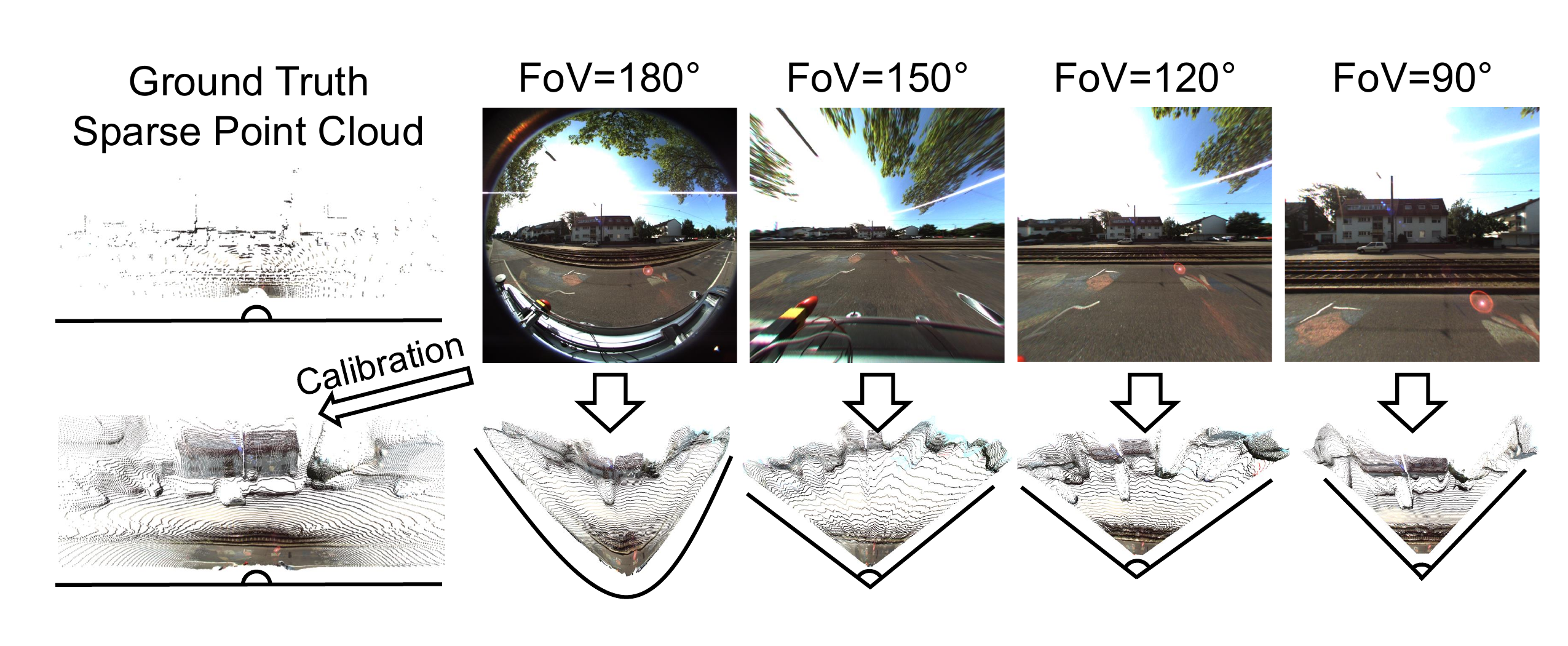}
    \vspace{-5mm}
    \caption{
        Foundation models for 3D reconstruction generally fail on fisheye images.
        \textbf{Top}: Ground truth sparse LiDAR points of a fisheye image, with undistorted perspective images of varying FoVs.
        \textbf{Bottom}: Reconstructed point clouds from the corresponding images by \cite{keetha2025mapanything}; the leftmost one is calibrated by our method.
        Naively undistorting the fisheye image for 3D reconstruction either sacrifices scene coverage or introduces extreme resampling artifacts and peripheral stretching that deviate from the model's training distribution, as evident from the mismatch of predicted and actual FoVs.
    }
    \label{fig:undistortion}
    \vspace{-5mm}
\end{figure*}

Generalist feed-forward foundation models \cite{wang2025vggt,wang2025pi,keetha2025mapanything} have recently emerged as a dominant paradigm in multi-view 3D reconstruction (3R) from uncalibrated images, unifying traditionally disparate geometric tasks (camera pose, depth, point map, and field-of-view estimation) into a single, cohesive framework. As they are trained on large-scale datasets of abundant perspective (rectilinear) images, these foundation models generalize well across 3D scenes. Yet, when evaluated on images captured by fisheye cameras, which are commonly used in spatial applications from robotics to Augmented/Virtual Reality (AR/VR) \cite{schops2019bad,caruso2015large,yeshwanth2023scannet++,liao2022kitti} due to their large scene coverage for situational awareness, these models produce degraded outcomes due to curvilinear distortions inherent to fisheye optics (\cref{fig:undistortion}). As many spatial applications are deployed on platforms with wide field-of-view (FoV) cameras or mixed-camera systems, e.g., comprising both perspective and fisheye cameras, these foundation models have been unsuitable for direct integration into these commercial systems.

A conventional approach to handling wide FoV images is to pre-process them through a rectification step, which transforms the fisheye images to a rectilinear pinhole viewport. Yet, this comes with a significant loss of peripheral coverage (due to cropping), which was the precise advantage that motivated the use of wide FoV cameras in the first place. On the other hand, retaining the original fisheye FoV in perspective images results in images with extreme resampling artifacts. Furthermore, the undistorted image would similarly degrade the performance of downstream models due to processing of ultra-wide-angle projections that lie far outside the natural distribution of their training data, which again leads to performance degradation as illustrated in \cref{fig:undistortion}. Additionally, the rectification process incurs latency and relies on accurate calibration, where errors accumulate and propagate through the vision system and further degrade estimates. 

An alternative approach is to train a model from scratch or finetune an existing model specifically for fisheye images; however, unlike the abundance of publicly available perspective image datasets, there is a comparatively small corpus of fisheye images, and even fewer fisheye images with ground truth to support 3D tasks. This fisheye data scarcity makes it challenging to train a foundation model that generalizes well across diverse 3D scenes. While finetuning may be an option, this risks model parameters drifting away from a desirable minimum, resulting in a loss of fidelity or generalizability. In both cases, the resulting model becomes camera‑specific, requiring separate models to be deployed for mixed‑camera systems, which in turn increases operational overhead.

Finally, one may also train a foundation model from scratch on a joint corpus of perspective and fisheye datasets. Not only is this computationally taxing for multi-view foundation models, but it also presents a high imbalance of data between the two camera types, resulting in trade-offs between fidelity and generalizability \cite{piccinelli2025unik3d}. We consider whether there exists a middle ground that enables a model to infer with similar fidelity on both perspective and fisheye images to avoid the operational overhead of deploying multiple models, but also without the computational overhead of large-scale training.

Given that existing foundation models already exhibit high generalizability across 3D scenes and that errors stem primarily from curvilinear distortions, we aim to recalibrate the fisheye input, not in the image space as done conventionally, but in the model's latent space, such that its embeddings become conducive to an existing foundation model pre-trained on perspective images. To this end, we propose \textit{Fisheye3R}, a lightweight and efficient adaptation framework that can potentially extend any pre-trained foundation model to fisheye imagery with similar fidelity as perspective images and without loss of generalizability.

Taking advantage of the transformer architectures \cite{dosovitskiy2020image} in foundation 3R models, we adopt a token-based adaptation strategy, where trainable calibration tokens are inserted into transformer blocks to extend the performance of a given pre-trained model to fisheye images. The influence of these tokens is controlled by a masked attention scheme, where their effect is eliminated for perspective images and activated on fisheye images, enabling the model to remain backwards compatible with perspective imagery. Considering the limited availability of ground truth for fisheye images on 3D tasks, we propose three distinct learning schemes to accommodate various degrees of data accessibility:
(i) self-supervised learning with unlabeled perspective RGB images only;
(ii) supervised learning with annotated perspective data;
and (iii) supervised learning with perspective and fisheye data.
Through extensive evaluation, we demonstrate that Fisheye3R consistently improves 3D reconstruction performance on fisheye images across all three schemes, providing a versatile solution regardless of data constraints.
Our work offers the following advantages:
\begin{itemize}
    \item Generalizability. We apply our framework generically to adapt three unified multi-view feed-forward foundation 3D reconstruction models \cite{wang2025vggt,wang2025pi,keetha2025mapanything}.
    \item Efficiency. We show that the models can be adapted through the insertion of some lightweight learnable tokens, requiring minimal training time and negligible computational overhead compared to full-parameter finetuning.
    \item Backwards compatibility. We demonstrate that the model can fully preserve the original performance on perspective data while gaining the unique capability to handle heterogeneous camera models within a single sequence.
    \item Data flexibility. Our method supports three different learning schemes tailored to varying data regimes, with which we observe consistent improvements on fisheye data, even with unlabeled perspective images for training.
\end{itemize}

\section{Related Work}
\label{sec:related_work}

\paragraphtitle{Traditional 3D Reconstruction.}
To recover 3D geometry from single or multiple views, traditional methodologies typically employ Structure-from-Motion (SfM) \cite{schonberger2016structure,pan2024global,gherardi2010improving,wu2013towards,sweeney2015optimizing,agarwal2011building,frahm2010building,snavely2006photo} to estimate sparse point clouds and camera parameters, followed by Multi-View Stereo (MVS)~\cite{schonberger2016pixelwise,yao2018mvsnet,gu2020cascade,ding2022transmvsnet,wang2021patchmatchnet,zhao2023mvpsnet} to reconstruct dense scene geometry.
These multi-stage pipelines rely on a sequence of independent tasks, including keypoint detection, feature extraction, matching, triangulation, and bundle adjustment, where optimization noise and errors can accumulate at each step.

\paragraphtitle{Feed-Forward 3D Reconstruction.}
Recently, transformer-based models have emerged to estimate 3D geometry from multiple images.
DUSt3R \cite{wang2024dust3r} regressed point maps from image pairs, while MASt3R \cite{leroy2024grounding} further improved upon this by adding local feature regression to enhance reconstruction accuracy.
Subsequent variants further extended their capabilities to dynamic scenes \cite{zhang2024monst3r}, streaming data \cite{wang2025continuous}, and parallelized reconstruction \cite{yang2025fast3r}.

\paragraphtitle{Unified 3D Feed-Forward Models.}
Unified 3D foundation models serve as general-purpose geometric backbones capable of solving multiple downstream tasks simultaneously.
VGGT \cite{wang2025vggt} unifies 3D vision by treating it as a sequence-to-sequence translation problem, regressing camera parameters, depth maps, point maps, and point tracks from uncalibrated images in a single forward pass.
$\pi^3$ \cite{wang2025pi} addresses VGGT's limitation regarding reference frame dependency by introducing a permutation-equivariant architecture, enabling robust prediction of local point maps and camera poses regardless of the input order.
MapAnything \cite{keetha2025mapanything} further expands the versatility by leveraging a multi-modal encoder to infer metric geometry from images and optional priors.

\paragraphtitle{Fisheye-Aware 3D Perception.}
Utilizing uncalibrated fisheye inputs for 3D reconstruction remains a challenging problem due to arbitrary distortions.
Traditional pipelines rely on integrating explicit calibration \cite{scaramuzza2006flexible,mei2007single,geyer2000unifying,kannala2006generic} into geometric solvers \cite{zhao2025fisheyedepth, matsuki2018omnidirectional, caruso2015large}, while recent learning-based attempts often circumvent native distortion by learning from canonical projections \cite{guo2025depth,li2021omnidirectional,wang2020360sd,zhao2025fastvidar,deng2025omnistereo}.
Other data-driven approaches focus on isolated subtasks: \cite{veicht2024geocalib,tirado2025anycalib} target intrinsic calibration, while \cite{lichy2024fova} adapts stereo depth via warping.
More recently, \cite{piccinelli2025unik3d} proposes a universal model for local camera 3D estimation using spherical basis functions, while \cite{gangopadhyay2025extending} employs calibration tokens and \cite{gangopadhyay2026from} proposes distortion extenders to adapt monocular depth estimation models to fisheye cameras.

\begin{figure*}[t]
    \centering
    \includegraphics[width=\linewidth]{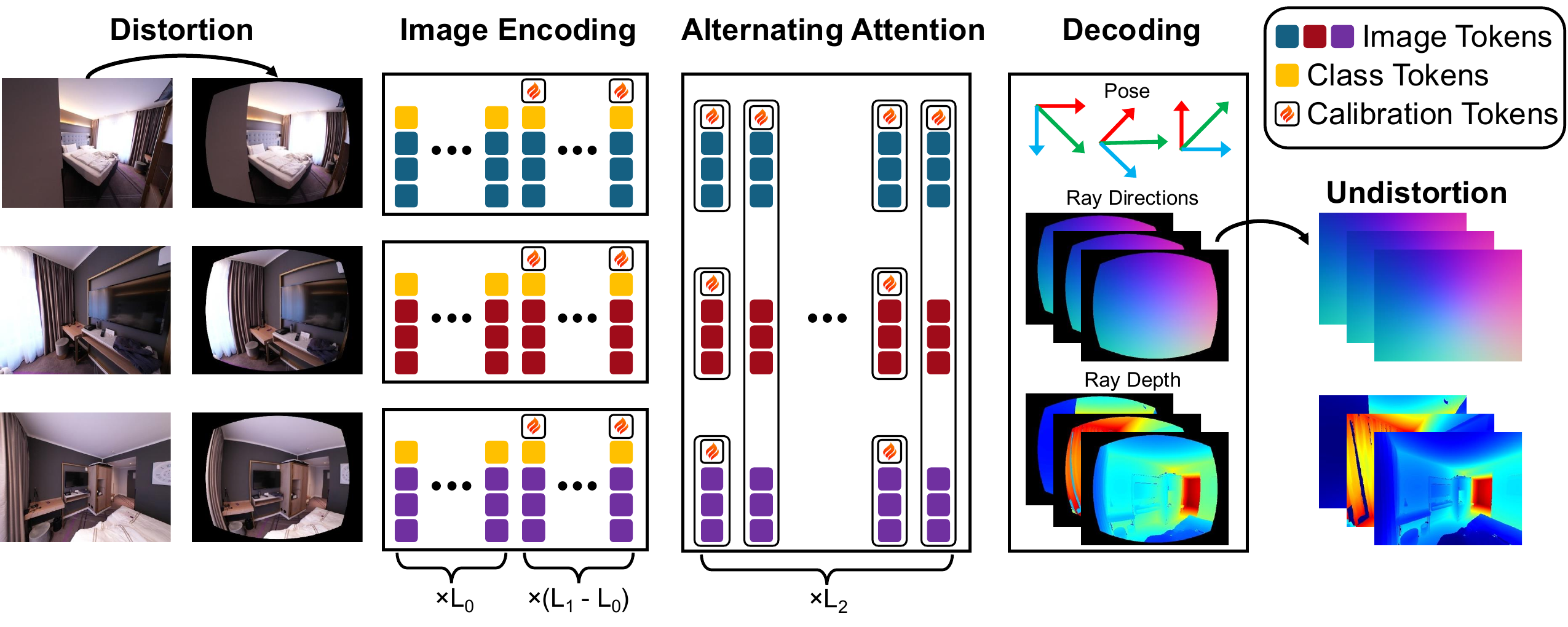}
    \vspace{-8mm}
    \caption{
        \textbf{Fisheye adaptation framework with perspective training data.}
        Our method leverages existing perspective datasets by synthesizing curvilinear distortion in the input sequences.
        The distorted images are processed by the frozen transformer backbone, where learnable calibration tokens are injected into the encoder layers. 
        The decoder predicts dense geometric attributes, which are undistorted for loss computation.
        Supervision is derived either from ground truth or through a self-supervised distillation of the model's own predictions on the original perspective sequence.
    }
    \label{fig:method}
    \vspace{-5mm}
\end{figure*}

\section{Method}
\label{sec:method}

\subsection{Problem Setup and Representations}
\label{sec:setup}

Multi-view feed-forward 3D reconstruction models \cite{wang2025vggt, wang2025pi, keetha2025mapanything} estimate camera pose and dense geometric attributes (e.g., depth, point maps, ray directions) from a set of uncalibrated RGB images capturing the same 3D scene.
This process generally consists of three stages.

\paragraphtitle{Image encoding.}
Consider a sequence of $S$ images $(I_s)_{s=1}^S$ from a scene where $I_s \in \R^{H \times W \times 3}$.
Each frame is independently passed to some pre-trained image encoder, typically DINOv2 \cite{oquab2023dinov2}, for feature extraction.
Specifically, for each $s \in [S] = \{1, \dots, S\}$, the image $I_s$ is divided into $N = HW/P^2$ patches of size $P \times P$ pixels, which are then projected to patch embeddings $\{\boldsymbol{x}_{s,n}^{(0)}\}_{n=1}^N$ where $\boldsymbol{x}_{s,n}^{(0)} \in \R^d$ and concatenated with a class token $\boldsymbol{x}_{s,0}^{(0)} = \boldsymbol{x}_\text{class} \in \R^d$ as
\begin{equation}
    \boldsymbol{x}_s^{(0)} = \left[
        \boldsymbol{x}_{s,0}^{(0)},
        \boldsymbol{x}_{s,1}^{(0)},
        \dots,
        \boldsymbol{x}_{s,N}^{(0)}
    \right] \in \R^{(1+N) \times d},
\end{equation}
where $[\cdot]$ denotes concatenation and $d$ is the embedding dimension.
For an encoder with $L_1$ layers, the encoder module $f_\text{E}^{(\ell)}$ at a layer $\ell \in [L_1]$ derives
\begin{equation}
    \label{eq:enc}
    \boldsymbol{x}_s^{(\ell)} = f_\text{E}^{(\ell)} \left( \boldsymbol{x}_s^{(\ell-1)} \right) \in \R^{(1+N) \times d}
\end{equation}
through multi-head self-attention and MLP blocks \cite{dosovitskiy2020image}.
After the last layer, the $N$ patch tokens are passed to the next stage as $\boldsymbol{z}_s^{(0)} = [ \boldsymbol{x}_{s,n}^{(L_1)} ]_{n=1}^N \in \R^{N \times d}$.

\paragraphtitle{Alternating attention.}
With a total of $L_2$ alternating attention (AA) blocks, each block $\ell \in [L_2]$ sequentially applies frame-wise and global self-attention as
\begin{align}
    \label{eq:frame}
    \boldsymbol{z}_s'^{(\ell)}
    &= f_\text{F}^{(\ell)} \left( \boldsymbol{z}_s^{(\ell-1)} \right) \in \R^{N \times d}, \; \forall s \in [S],
    \\
    \label{eq:global}
    \left[ \boldsymbol{z}_s^{(\ell)} \right]_{s=1}^S
    &= f_\text{G}^{(\ell)} \left( \left[ \boldsymbol{z}_s'^{(\ell)} \right]_{s=1}^S \right) \in \R^{SN \times d},
\end{align}
where the tokens of each frame attend to each other in frame-wise attention layer $f_\text{F}^{(\ell)}$ and the tokens across all frames attend to each other in global attention layer $f_\text{G}^{(\ell)}$.
The order of frame and global attention can vary depending on the model's architectural design.

\paragraphtitle{Scene representation prediction.}
The outputs of the AA module $\{ \boldsymbol{z}_s^{(L_2)} \}_{s=1}^S$ are passed to decoder $f_\text{D}$ as
\begin{equation}
    E_s, D_s = f_\text{D}\left( \boldsymbol{z}_s^{(L_2)} \right),
\end{equation}
where $E_s = \left[ R_{s} | \boldsymbol{t}_s \right]$ denotes camera pose for frame $s$ with rotation $R_{s} \in \R^{3 \times 3}$ and translation $\boldsymbol{t}_s \in \R^{3}$, and $D_s \in \R^{H \times W \times C}$ is the combination of dense predictions whose types vary across different architectures:
VGGT \cite{wang2025vggt} predicts depth map in the local camera coordinates and point map in the global world coordinates;
$\pi^3$ \cite{wang2025pi} predicts a local point map and infers the global point map from the local one and the predicted camera pose;
MapAnything \cite{keetha2025mapanything} predicts local ray directions and ray depths to derive local camera coordinates, from which global point coordinates are inferred with the predicted pose.

\subsection{Model Adaptation with Calibration Tokens}

While foundation models trained on massive rectilinear datasets excel at processing sequences of perspective images, they fail to generalize to fisheye imagery.
When presented with fisheye inputs, the frozen backbone generates features that fall outside the learned data manifold of perspective images, resulting in distorted geometric reconstructions and erroneous camera pose estimates.

To bridge this gap, we align fisheye feature representations with those of perspective images by introducing $K$ learnable calibration tokens into each transformer layer. 
As illustrated in \cref{fig:method}, these tokens are inserted into all the encoder layers except the initial $L_0$ layers to modulate the high-level features.
The learnable tokens ``recalibrate'' the values of fisheye features such that they resemble those of perspective images to enable a pre-trained model to infer attributes of the 3D scene with high fidelity.
Specifically, we reformulate the transformer encoder, frame-wise, and global attention layers from \cref{eq:enc,eq:frame,eq:global} as
\begin{align}
    \left[
        \boldsymbol{x}_s^{(\ell)},
        \left[ \boldsymbol{\phi}_{\text{E},k}'^{(\ell)} \right]_{k=1}^K
    \right]
    &= f_\text{E}^{(\ell)} \left(\left[
        \boldsymbol{x}_s^{(\ell-1)},
        \left[ \boldsymbol{\phi}_{\text{E},k}^{(\ell)} \right]_{k=1}^K
    \right]\right), \; \forall \ell \in (L_0, L_1],
    \\
    \left[
        \boldsymbol{z}_s'^{(\ell)},
        \left[ \boldsymbol{\phi}_{\text{F},k}'^{(\ell)} \right]_{k=1}^K
    \right]
    &= f_\text{F}^{(\ell)} \left(\left[
        \boldsymbol{z}_s^{(\ell-1)},
        \left[ \boldsymbol{\phi}_{\text{F},k}^{(\ell)} \right]_{k=1}^K
    \right]\right), \; \forall \ell \in [L_2],
    \\
    \left[
        \left[ \boldsymbol{z}_s^{(\ell)} \right]_{s=1}^S,
        \left[ \boldsymbol{\phi}_{\text{G},k}'^{(\ell)} \right]_{k=1}^K
    \right]
    &= f_\text{G}^{(\ell)} \left(\left[
        \left[\boldsymbol{z}_s'^{(\ell)} \right]_{s=1}^S,
        \left[ \boldsymbol{\phi}_{\text{G},k}^{(\ell)} \right]_{k=1}^K
    \right]\right), \; \forall \ell \in [L_2],
\end{align}
where $\boldsymbol{\phi}_{\text{M},k}^{(\ell)} \in \R^d$ represents the $k$-th calibration token at the $\ell$-th layer of module $\text{M} \in \{\text{E}, \text{F}, \text{G}\}$.
In other words, at each encoder and frame-wise attention layer, $K$ calibration tokens adapt the $N$ image tokens separately.
At each global attention layer, $K$ calibration tokens adapt the $SN$ image tokens of the entire sequence jointly.
Each layer has its own set of calibration tokens, which are dropped immediately at each layer to localize the latent calibration effect to each layer.
During training, only calibration tokens are learned, while the original backbone is entirely frozen for efficiency and to preserve the model's knowledge on perspective images.

\subsection{Learning}
\label{sec:learning}

We denote the entire feed-forward model by $f$. With a perspective image sequence $I^\text{p} = (I_s^\text{p})_{s=1}^S$, we predict
\begin{equation}
    \hat{E}^\text{p}, \hat{D}^\text{p} = f\left( I^\text{p} \right),
\end{equation}
where $\hat{E}^\text{p}$ and $\hat{D}^\text{p}$ denote the predicted camera poses and dense geometry outputs for the perspective sequence.
To adapt the model to fisheye data, we can use a sequence of fisheye images $I^\text{f} = (I_s^\text{f})_{s=1}^S$ for training, i.e.,
\begin{equation}
    \hat{E}^\text{f}, \hat{D}^\text{f} = f\left( I^\text{f}, \boldsymbol{\phi} \right),
\end{equation}
where $\boldsymbol{\phi}$ is the set of all calibration tokens and $\hat{E}^\text{f}, \hat{D}^\text{f}$ are outputs corresponding to the fisheye sequence.

The ideal scenario assumes abundant fisheye data is available for training; given this is not the case in practice, we opt for an alternative of generating fisheye images from perspective ones.
Particularly, we adopt Kannala-Brandt distortion model \cite{kannala2006generic} for fisheye image synthesis from perspective images: $I_s^\text{f} = T(I_s^\text{p})$ for each frame $s$ where $T$ denotes the distortion transformation from perspective to fisheye image.
The inverse distortion operation can be represented by $T^{-1}$, which is used to obtain undistorted geometric predictions from fisheye predictions as $\hat{D}_s^\text{p} = T^{-1}(\hat{D}_s^\text{f})$.
The camera extrinsic parameters are unrelated to the camera type, thus they do not need to be transformed: $\hat{E}_s^\text{p} = \hat{E}_s^\text{f}$.

Finally, the calibration tokens can be learned by supervising the outputs. The supervision type depends on the data availability.
We propose three supervision schemes according to the types of data available.

\begin{figure*}[t]
    \centering
    \includegraphics[width=\linewidth]{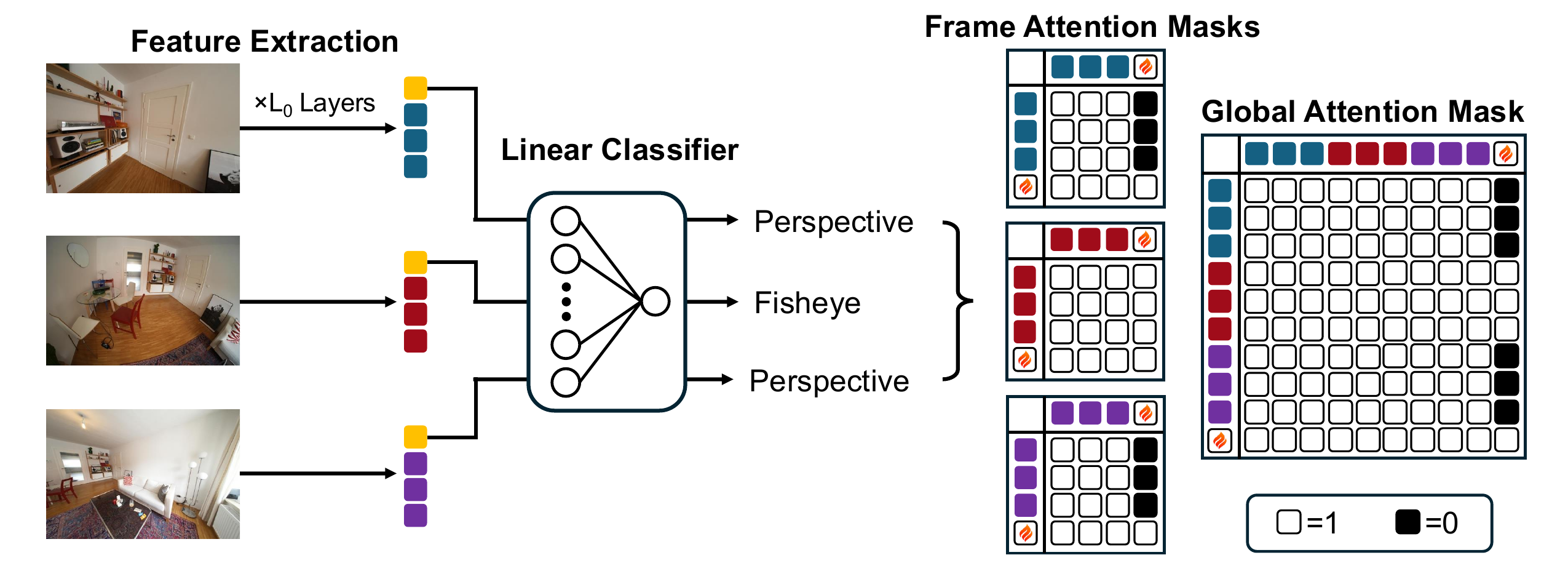}
    \vspace{-6mm}
    \caption{
        \textbf{Masked attention for mixture sequences.}
        To handle a sequence of mixed camera types, we introduce a masked attention mechanism to block the influence of calibration tokens on perspective image tokens.
        After $L_0$ layers of feature extraction in image encoder, the class tokens are passed to a linear classifier for camera type prediction.
        Then, frame-wise and global attention masks are created accordingly and used in all the remaining encoder layers, including the $L_1 - L_0$ ones in the image encoder, and the $L_2$ frame and $L_2$ global attention layers in alternating attention.
    }
    \label{fig:method_masked_attention}
    \vspace{-5mm}
\end{figure*}

\paragraphtitle{Self-supervised learning with perspective images (SSL).}
If only perspective RGB images $I^\text{p}$ are available with no ground truth, we generate pseudo-labels using the original model's predictions for self-supervision. Inspired by AugUndo \cite{wu2024augundo}, the loss function can therefore be expressed as
\begin{equation}
    \mathcal{L}_\text{SSL}^\text{p} = \mathcal{L} \left(f(I^\text{p}) ; T^{-1} \!\circ\! f(T(I^\text{p}), \boldsymbol{\phi}) \right).
    \label{eq:loss_ssl}
\end{equation}

\paragraphtitle{Supervised learning with perspective images (SL).}
If both perspective images $I^\text{p}$ and their ground truth $(E^\text{p}, D^\text{p})$ are available, they can be directly used for supervision as
\begin{equation}
    \mathcal{L}_\text{SL}^\text{p} = \mathcal{L} \left((E^\text{p}, D^\text{p}); T^{-1} \!\circ\! f(T(I^\text{p}), \boldsymbol{\phi}) \right).
    \label{eq:loss_sl}
\end{equation}

\paragraphtitle{Supervised learning with fisheye images (SL+).}
If fisheye images $I^\text{f}$ with ground truth $(E^\text{f}, D^\text{f})$ are available, we can train the calibration tokens via
\begin{equation}
    \mathcal{L}_\text{SL}^\text{f} = \mathcal{L} \left((E^\text{f}, D^\text{f}); f(I^\text{f}, \boldsymbol{\phi}) \right).
    \label{eq:loss_slp}
\end{equation}
The loss function $\mathcal{L}$ is the loss used by the original 3D reconstruction method.
Undistortion $T^{-1}$ is applied to the dense geometry outputs but not the pose.
Supervision is consistently applied at the spatial resolution of the ground truth observation, preventing artifacts that may arise from resampling.

\subsection{Controlled Adaptation with Masked Attention}

We further extend the model to handle unknown and potentially mixed types of cameras in a sequence during inference.
Though calibration tokens can learn to align fisheye features to perspective ones effectively, we observe that they may not be backwards compatible with perspective images once trained, i.e., given perspective images, the calibration tokens are likely to modulate their features, leading to performance degradation on perspective data.
This can be explained by the increased task complexity for calibration tokens to not only calibrate fisheye features but also preserve perspective ones simultaneously.
Therefore, we propose to control the effect of calibration tokens depending on the camera type of images observed by using a masked attention scheme (\cref{fig:method_masked_attention}).
The effect of calibration tokens is eliminated by a binary mask for perspective images, but is activated for fisheye images. This mechanism ensures backwards compatibility with perspective images and enables inference on mixed-camera sequences.

\paragraphtitle{Camera type classification.}
Due to the significant difference between perspective and fisheye images in terms of distortion characteristics, classifying the two types of images from their feature vectors is feasible.
As we have the class tokens readily available from the image encoder as a compact representation of the image feature, we employ them to determine the camera type of an image.
Formally, for a frame $I_s$, we extract its class token $\boldsymbol{x}_{s,0}^{(L_0)}$ from the $L_0$-th layer and pass it to an image classifier $\psi: \R^d \rightarrow (0, 1)$, which can be as simple as a linear classifier with sigmoid activation.
Consequently, a binary predicted camera type can be obtained for each frame $s$ as
\begin{equation}
    M_s = \mathbb{I}\left(
        \psi \left(\boldsymbol{x}_{s,0}^{(L_0)} \right) > 0.5
    \right) \in \{0, 1\},
\end{equation}
where $\mathbb{I}$ denotes the indicator function and $M_s$ is $1$ for fisheye and $0$ for perspective.

\begin{table*}[t]
    \centering
    \resizebox{\textwidth}{!}{
        \newcolumntype{M}{>{\centering\arraybackslash}m{9mm}}
        \begin{tabular}{cc | *{4}{MMM|} MMM }
            \toprule[1pt]
            \multicolumn{2}{c|}{\multirow{2}{*}{Dataset \& Method}} &
            \multicolumn{3}{c|}{Pose (Angular)} &
            \multicolumn{3}{c|}{Pose (Distance)} &
            \multicolumn{3}{c|}{Depth Map} &
            \multicolumn{3}{c|}{Point Map} &
            \multicolumn{3}{c}{FoV Map}
            \\
            \cline{3-5} \cline{6-8} \cline{9-11} \cline{12-14} \cline{15-17}
            & & RRA$\uparrow$ & RTA$\uparrow$ & AUC$\uparrow$
            & ATE$\downarrow$ & RPEt$\downarrow$ & RPEr$\downarrow$
            & Rel$\downarrow$ & RMSE$\downarrow$ & $\delta_1$$\uparrow$
            & Acc$\downarrow$ & Comp$\downarrow$ & CD$\downarrow$
            & hErr$\downarrow$ & vErr$\downarrow$ & AUC$\uparrow$
            \\ \midrule
            \multirow{12}{*}{\rotatebox{90}{ScanNet++ \cite{yeshwanth2023scannet++}}}
            
            & VGGT \cite{wang2025vggt}
            &{0.956}	&{0.759}	&{0.381}	&{0.200}	&{0.420}	&{13.877}	&{0.287}	&{0.440}	&{0.618}	&{0.122}	&{0.102}	&{0.112}	&{12.962}	&{10.253}	&{0.305}	
			
            \\
            & w/ \textbf{SSL}
            &{\textbf{0.999}}	&{0.813}	&\orange{0.533}	&\yellow{0.148}	&\orange{0.284}	&\red{6.097}	&\yellow{0.234}	&\yellow{0.354}	&\yellow{0.744}	&\orange{0.077}	&\orange{0.052}	&\orange{0.065}	&\orange{7.092}	&\orange{5.708}	&\red{0.605}	
			
            \\
            & w/ \textbf{SL}
            &{\textbf{0.999}}	&\yellow{0.910}	&\red{0.656}	&\red{0.097}	&\red{0.190}	&\red{4.855}	&\yellow{0.230}	&\yellow{0.343}	&\yellow{0.742}	&\orange{0.074}	&\red{0.046}	&\orange{0.060}	&\red{4.439}	&\red{3.614}	&\red{0.722}
			
            \\
            & w/ \textbf{SL+}
            &{\textbf{0.999}}	&\yellow{0.916}	&\red{0.677}	&\red{0.088}	&\red{0.168}	&\red{4.363}	&\yellow{0.222}	&\yellow{0.327}	&\yellow{0.759}	&\orange{0.073}	&\red{0.046}	&\orange{0.060}	&\red{4.602}	&\red{3.606}	&\red{0.718}	

            \\

            \cline{2-17}
            & $\pi^3$ \cite{wang2025pi}
            &{0.989}	&{0.814}	&{0.463}	&{0.164}	&{0.347}	&{10.651}	&{0.282}	&{0.421}	&{0.661}	&{0.095}	&{0.073}	&{0.084}	&{8.288}	&{6.332}	&{0.551}	
			
            \\
            & w/ \textbf{SSL}
            &{\textbf{0.999}}	&{0.844}	&\yellow{0.571}	&\orange{0.114}	&\orange{0.220}	&\orange{6.146}	&\yellow{0.239}	&\yellow{0.357}	&\yellow{0.741}	&\yellow{0.070}	&\orange{0.050}	&\yellow{0.060}	&\red{3.926}	&\red{3.090}	&\orange{0.773}	
			
            \\
            & w/ \textbf{SL}
            &{\textbf{0.999}}	&{0.872}	&\orange{0.617}	&\orange{0.101}	&\orange{0.194}	&\orange{5.424}	&\yellow{0.239}	&\yellow{0.350}	&\yellow{0.744}	&\yellow{0.068}	&\orange{0.048}	&\orange{0.058}	&\red{3.036}	&\red{2.561}	&\orange{0.810}	
			
            \\
            & w/ \textbf{SL+}
            &{\textbf{0.999}}	&\yellow{0.926}	&\red{0.719}	&\red{0.072}	&\red{0.139}	&\red{\textbf{3.710}}	&\yellow{0.212}	&\yellow{0.320}	&\yellow{0.771}	&\orange{0.064}	&\orange{\textbf{0.044}}	&\orange{0.054}	&\red{\textbf{2.941}}	&\red{3.018}	&\red{0.873}	
			
            \\
            
            \cline{2-17}
            & MapAnything \cite{keetha2025mapanything}
            &{0.993}	&{0.848}	&{0.519}	&{0.163}	&{0.310}	&{8.851}	&{0.274}	&{0.394}	&{0.672}	&{0.091}	&{0.076}	&{0.083}	&{6.631}	&{4.821}	&{0.647}	
            
            \\
            & w/ \textbf{SSL}
            &{\textbf{0.999}}	&\yellow{\textbf{0.972}}	&\orange{0.766}	&\red{0.070}	&\red{0.143}	&\orange{4.569}	&\orange{0.190}	&\yellow{0.283}	&\yellow{0.785}	&\orange{\textbf{0.057}}	&\orange{0.050}	&\orange{0.054}	&\red{3.146}	&\orange{2.575}	&\orange{0.875}	
			
            \\
            & w/ \textbf{SL}
            &{\textbf{0.999}}	&\yellow{0.959}	&\orange{0.730}	&\red{0.080}	&\orange{0.164}	&\orange{5.032}	&\yellow{0.195}	&\yellow{0.289}	&\yellow{0.786}	&\orange{0.059}	&\orange{0.050}	&\orange{0.054}	&\orange{4.011}	&\orange{3.049}	&\orange{0.853}	
			
            \\
            & w/ \textbf{SL+}
			&{\textbf{0.999}}	&\yellow{0.970}	&\orange{\textbf{0.774}}	&\red{\textbf{0.066}}	&\red{\textbf{0.134}}	&\red{4.370}	&\orange{\textbf{0.171}}	&\orange{\textbf{0.263}}	&\yellow{\textbf{0.810}}	&\orange{0.058}	&\orange{0.045}	&\orange{\textbf{0.051}}	&\red{3.264}	&\red{\textbf{2.074}}	&\orange{\textbf{0.927}}
            
            \\
            
            \midrule
            \multirow{12}{*}{\rotatebox{90}{ADT \cite{pan2023aria}}}
            & VGGT \cite{wang2025vggt}
            &{0.964}	&{0.777}	&{0.458}	&{0.134}	&{0.260}	&{10.055}	&{0.189}	&{0.739}	&{0.732}	&{0.237}	&{0.179}	&{0.208}	&{7.382}	&{7.566}	&{0.475}

            \\
            & w/ \textbf{SSL}
            &{0.997}	&\yellow{0.882}	&\orange{0.662}	&\orange{0.079}	&\orange{0.150}	&\red{2.761}	&\orange{0.110}	&\orange{0.502}	&\yellow{0.885}	&\orange{0.140}	&\orange{0.098}	&\orange{0.119}	&\red{1.992}	&\red{1.992}	&\red{0.983}

            \\
            & w/ \textbf{SL}
            &{0.995}	&\yellow{0.931}	&\red{0.746}	&\red{0.048}	&\red{0.095}	&\red{2.242}	&\red{0.082}	&\red{0.354}	&\yellow{0.930}	&\red{0.080}	&\red{0.053}	&\red{0.066}	&\red{1.418}	&\red{1.344}	&\red{\textbf{1.000}}

            \\
            & w/ \textbf{SL+}
            &{0.998}	&\yellow{0.948}	&\red{0.778}	&\red{0.038}	&\red{0.073}	&\red{\textbf{1.820}}	&\red{0.076}	&\red{0.334}	&\yellow{0.934}	&\red{0.072}	&\red{0.045}	&\red{0.059}	&\red{1.665}	&\red{1.598}	&\red{\textbf{1.000}}

            \\

            \cline{2-17}
            & $\pi^3$ \cite{wang2025pi}
            &{0.984}	&{0.893}	&{0.601}	&{0.078}	&{0.157}	&{7.148}	&{0.111}	&{0.410}	&{0.909}	&{0.126}	&{0.106}	&{0.116}	&{4.044}	&{3.961}	&{0.927}

            \\
            & w/ \textbf{SSL}
            &{\textbf{1.000}}	&{0.953}	&\yellow{0.776}	&\red{0.035}	&\red{0.069}	&\red{2.241}	&\orange{0.075}	&\yellow{0.332}	&{0.937}	&\orange{0.075}	&\orange{0.054}	&\orange{0.064}	&\red{1.217}	&\red{\textbf{0.989}}	&{\textbf{1.000}}

            \\
            & w/ \textbf{SL}
            &{0.996}	&{0.945}	&\yellow{0.768}	&\orange{0.041}	&\orange{0.079}	&\red{3.325}	&\orange{\textbf{0.073}}	&\yellow{\textbf{0.323}}	&{\textbf{0.940}}	&\orange{\textbf{0.068}}	&\red{0.047}	&\red{0.058}	&\red{\textbf{1.161}}	&\red{1.068}	&{\textbf{1.000}}

            \\
            & w/ \textbf{SL+}
            &{0.996}	&{\textbf{0.960}}	&\orange{\textbf{0.819}}	&\red{\textbf{0.034}}	&\red{\textbf{0.066}}	&\red{2.799}	&\yellow{0.087}	&\yellow{0.362}	&{0.915}	&\orange{0.074}	&\red{\textbf{0.039}}	&\red{\textbf{0.056}}	&\red{1.211}	&\red{1.534}	&{0.997}

            \\

            \cline{2-17}
            & MapAnything \cite{keetha2025mapanything}
            &{0.959}	&{0.793}	&{0.448}	&{0.125}	&{0.264}	&{9.601}	&{0.145}	&{0.411}	&{0.860}	&{0.164}	&{0.130}	&{0.147}	&{8.313}	&{8.141}	&{0.399}

            \\
            & w/ \textbf{SSL}
            &{0.996}	&\yellow{0.908}	&\red{0.704}	&\red{0.060}	&\red{0.126}	&\red{3.525}	&\orange{0.090}	&\yellow{0.333}	&{0.921}	&\red{0.079}	&\red{0.056}	&\red{0.068}	&\red{2.032}	&\red{2.015}	&\red{\textbf{1.000}}

            \\
            & w/ \textbf{SL}
            &{0.997}	&\yellow{0.903}	&\red{0.691}	&\orange{0.063}	&\orange{0.134}	&\red{3.421}	&\orange{0.091}	&\yellow{0.337}	&{0.920}	&\red{0.081}	&\red{0.054}	&\red{0.068}	&\red{1.987}	&\red{2.001}	&\red{\textbf{1.000}}

            \\
            & w/ \textbf{SL+}
            &{\textbf{1.000}}	&\yellow{0.934}	&\red{0.753}	&\red{0.045}	&\red{0.080}	&\red{2.111}	&\orange{0.087}	&\yellow{0.330}	&{0.925}	&\red{0.072}	&\red{0.043}	&\red{0.058}	&\red{1.338}	&\red{1.101}	&\red{\textbf{1.000}}
            
            \\
            
            \midrule
            \multirow{12}{*}{\rotatebox{90}{KITTI360 \cite{liao2022kitti}}}
            & VGGT \cite{wang2025vggt}
            &{0.829}	&{0.968}	&{0.440}	&{0.768}	&{2.624}	&{20.242}	&{0.270}	&{6.736}	&{0.565}	&{1.029}	&{1.580}	&{1.304}	&{19.430}	&{7.273}	&{0.212}

            \\
            & w/ \textbf{SSL}
            &\yellow{0.945}	&{0.989}	&\orange{0.592}	&\orange{0.537}	&\orange{1.814}	&\orange{12.882}	&\yellow{0.200}	&\yellow{5.678}	&\yellow{0.724}	&{0.956}	&\yellow{1.249}	&\yellow{1.102}	&\yellow{15.737}	&\yellow{5.880}	&\orange{0.313}

            \\
            & w/ \textbf{SL}
            &\yellow{0.982}	&{0.991}	&\orange{0.649}	&\orange{0.443}	&\orange{1.503}	&\orange{10.290}	&\yellow{0.204}	&\orange{4.456}	&\yellow{0.694}	&\yellow{0.893}	&\yellow{1.148}	&\yellow{1.021}	&\orange{11.953}	&\orange{4.247}	&\red{0.435}

            \\
            & w/ \textbf{SL+}
			&\yellow{\textbf{1.000}}	&{\textbf{0.993}}	&\red{0.904}	&\red{0.117}	&\red{0.356}	&\red{2.409}	&\red{0.111}	&\orange{3.955}	&\red{0.890}	&\orange{0.644}	&\orange{0.808}	&\orange{0.726}	&\red{4.790}	&\red{2.252}	&\red{0.757}

            \\
            
            \cline{2-17}
            & $\pi^3$ \cite{wang2025pi}
			&{0.955}	&{0.961}	&{0.548}	&{0.428}	&{1.840}	&{11.717}	&{0.153}	&{4.159}	&{0.806}	&{0.772}	&{1.081}	&{0.927}	&{11.731}	&{4.316}	&{0.444}

            \\
            & w/ \textbf{SSL}
            &{0.981}	&{0.986}	&\yellow{0.649}	&\yellow{0.365}	&\yellow{1.457}	&\yellow{9.938}	&\orange{0.101}	&{3.923}	&\yellow{0.897}	&{0.758}	&\yellow{0.895}	&\yellow{0.827}	&\yellow{8.864}	&\yellow{3.804}	&\yellow{0.531}

            \\
            & w/ \textbf{SL}
            &{0.965}	&{\textbf{0.993}}	&\yellow{0.634}	&{0.404}	&\yellow{1.531}	&{11.555}	&\orange{0.105}	&{4.091}	&\yellow{0.893}	&{0.749}	&\yellow{0.909}	&\yellow{0.829}	&\orange{7.816}	&\yellow{3.203}	&\yellow{0.566}

            \\
            & w/ \textbf{SL+}
            &{\textbf{1.000}}	&{\textbf{0.993}}	&\red{\textbf{0.942}}	&\red{\textbf{0.092}}	&\red{\textbf{0.239}}	&\red{\textbf{1.484}}	&\orange{0.097}	&\yellow{3.645}	&\yellow{\textbf{0.922}}	&\orange{\textbf{0.399}}	&\red{\textbf{0.475}}	&\red{\textbf{0.437}}	&\orange{6.316}	&{4.001}	&\red{0.751}

            \\
            
            \cline{2-17}
            & MapAnything \cite{keetha2025mapanything}
			&{0.818}	&{0.947}	&{0.428}	&{1.215}	&{3.023}	&{20.568}	&{0.258}	&{5.775}	&{0.607}	&{1.097}	&{1.505}	&{1.301}	&{18.034}	&{6.642}	&{0.259}

            \\
            & w/ \textbf{SSL}
            &\yellow{0.906}	&{0.989}	&\yellow{0.540}	&\orange{0.707}	&\yellow{2.152}	&\yellow{15.528}	&\orange{0.156}	&\yellow{4.294}	&\orange{0.809}	&\yellow{0.933}	&\yellow{1.098}	&\yellow{1.015}	&\yellow{15.739}	&{6.025}	&\yellow{0.291}

            \\
            & w/ \textbf{SL}
			&\yellow{0.906}	&{0.982}	&\yellow{0.538}	&\orange{0.682}	&\yellow{2.141}	&\yellow{15.342}	&\orange{0.153}	&\yellow{4.458}	&\orange{0.814}	&\yellow{0.938}	&\yellow{1.094}	&\yellow{1.016}	&\yellow{15.750}	&{5.998}	&\yellow{0.295}

            \\
            & w/ \textbf{SL+}
			&\yellow{\textbf{1.000}}	&{0.992}	&\red{0.917}	&\red{0.152}	&\red{0.350}	&\red{1.565}	&\red{\textbf{0.091}}	&\orange{\textbf{3.282}}	&\red{\textbf{0.922}}	&\red{0.549}	&\red{0.601}	&\red{0.575}	&\red{\textbf{2.963}}	&\red{\textbf{1.938}}	&\red{\textbf{0.805}}
            
            \\
            \bottomrule[1pt]
        \end{tabular}
    }
    \vspace{1mm}
    \caption{
        Quantitative results of fisheye adaptation on three models, three datasets, and three learning schemes.
        Compared to the baselines, the cell colors denote an improvement of \highlightinline{tabyellow}{10\%-30\%}, \highlightinline{taborange}{30\%-50\%}, and \highlightinline{tabred}{50\%+}.
        The best numbers are marked in bold.
    }
    \label{tab:quantitative}
    \vspace{-5mm}
\end{table*}

\paragraphtitle{Attention masking.}
With such binary camera type predictions, the frame-wise attention mask $M_{\text{F},s}$ for each frame $s$ and the global attention mask $M_\text{G}$ for the sequence can be respectively constructed as
\begin{equation}
    M_{\text{F},s} = \begin{pmatrix}
        \mathbf{1}_{N \times N} & M_s \cdot \mathbf{1}_{N \times K}
        \\
        \mathbf{1}_{K \times N} & \mathbf{1}_{K \times K}
    \end{pmatrix},
    \qquad
    M_\text{G} = \begin{pmatrix}
        \mathbf{1}_{SN \times SN} & \begin{pmatrix}
            M_1 \cdot \mathbf{1}_{N \times K}
            \\
            M_2 \cdot \mathbf{1}_{N \times K}
            \\
            \vdots
            \\
            M_S \cdot \mathbf{1}_{N \times K}
        \end{pmatrix}
        \\
        \mathbf{1}_{K \times SN} & \mathbf{1}_{K \times K}
    \end{pmatrix},
\end{equation}
where $N$ is the number of image tokens per frame, $K$ is the number of calibration tokens per layer, and $\mathbf{1}$ represents a matrix of ones.
The $(i, j)$-th entry of each mask controls whether the $j$-th input token affects the $i$-th output token.
Then, $M_{\text{F},s}$ are passed to the remaining $L_1 - L_0$ image encoder layers and $L_2$ frame-wise layers in the AA module, while $M_\text{G}$ is passed to the global attention layers in the AA module.
Altogether, these masks nullify the effect of calibration tokens on perspective frames.

\begin{figure*}[t!]
    \centering
    \includegraphics[width=\linewidth]{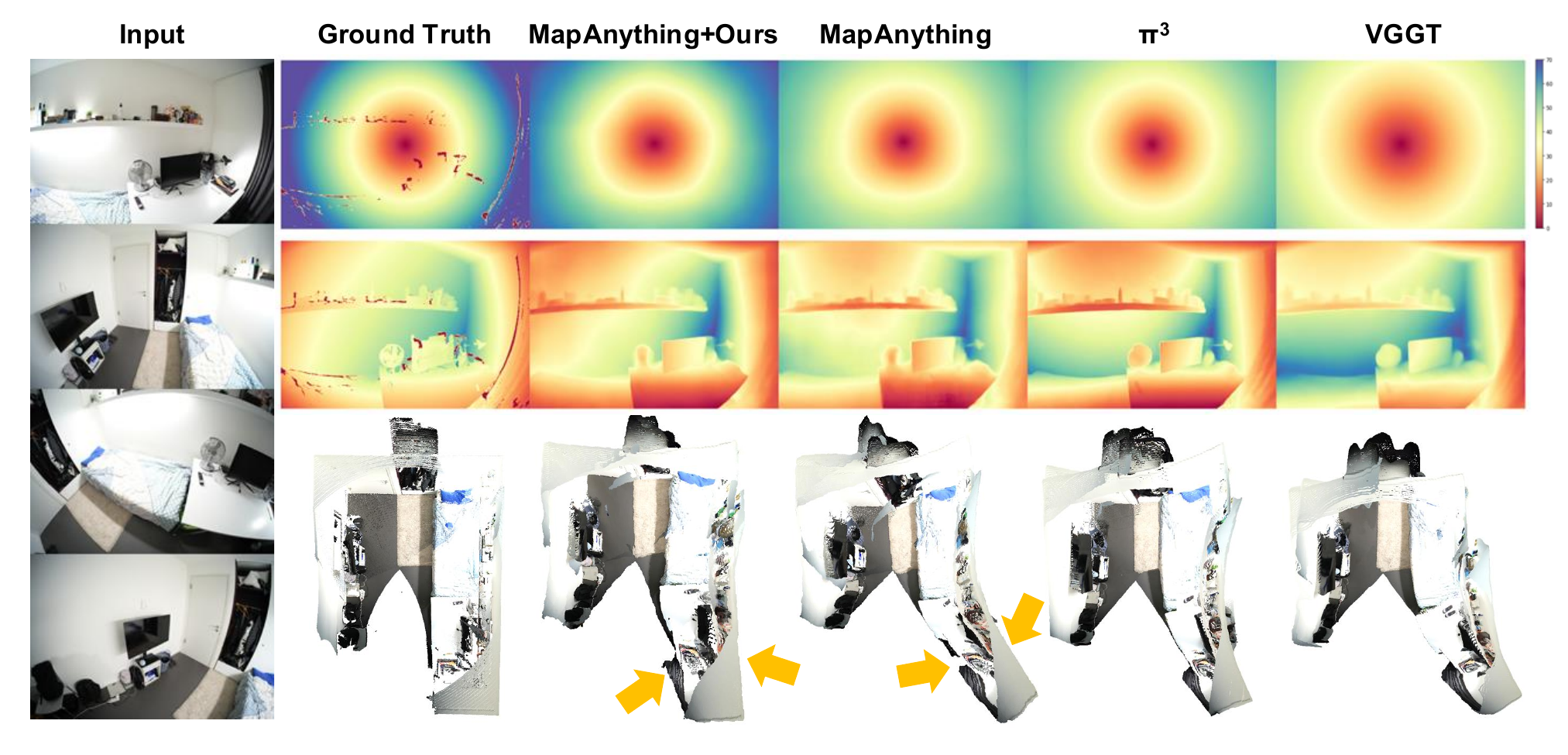}
    \vspace{-5mm}
    \includegraphics[width=\linewidth]{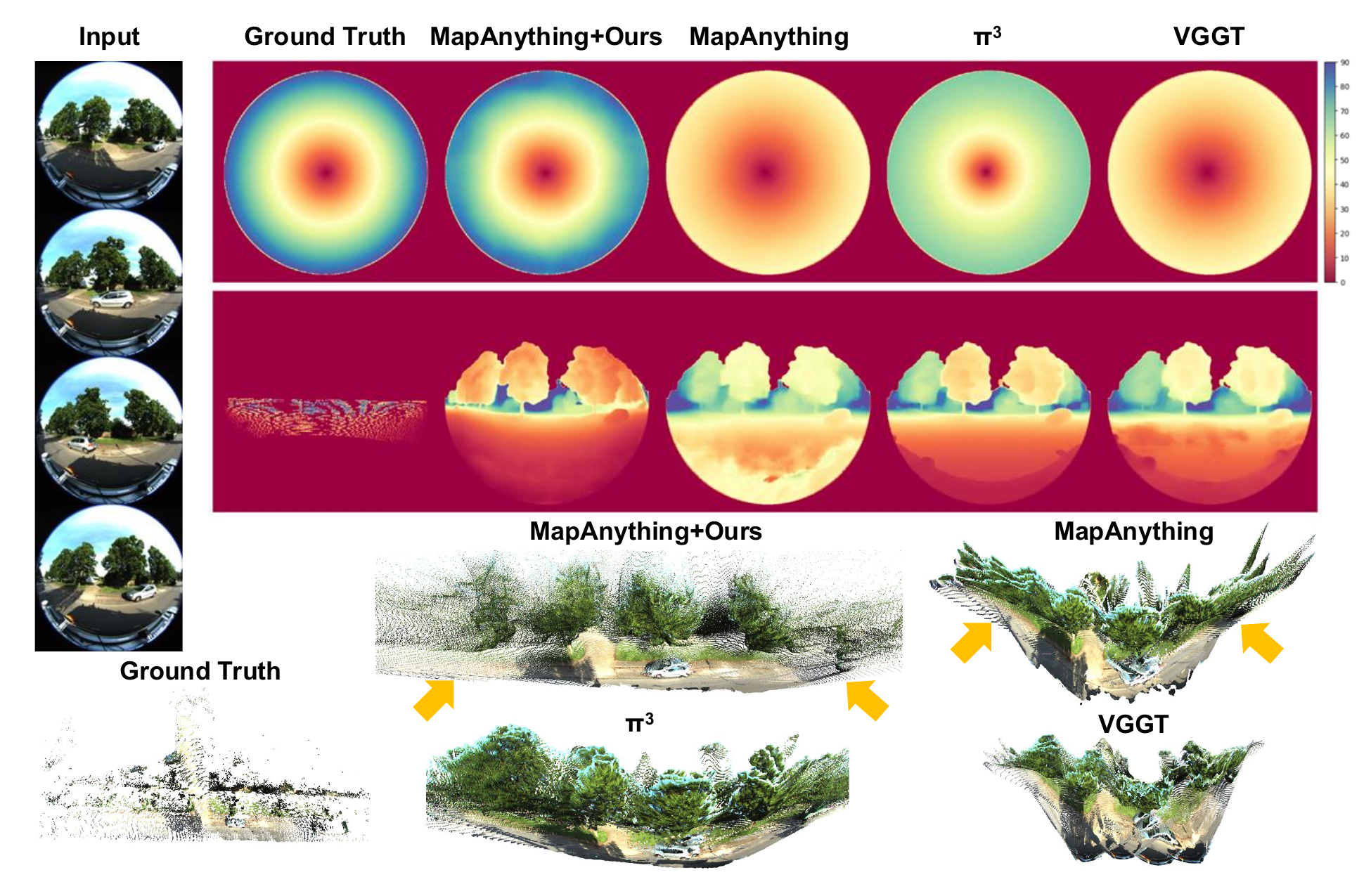}
    \caption{
        Qualitative comparison of FoV, depth, and point map estimation on an indoor (top) and outdoor (bottom) fisheye sequence.
        We apply calibration tokens to MapAnything and consistently produce more geometrically consistent reconstructions compared to the baseline models.
        The yellow arrows highlight the regions where our calibration tokens effectively rectify significant geometric distortions.
    }
    \label{fig:qualitative}
    \vspace{-2mm}
\end{figure*}

\section{Experiments}
\label{sec:experiments}

\subsection{Settings}
\label{sec:settings}

\paragraphtitle{Datasets.}
We train the models with 3 learning schemes as mentioned in \cref{sec:learning}.
For \textbf{SSL} and \textbf{SL}, we use 6 perspective datasets for training: ScanNet++ (perspective) \cite{yeshwanth2023scannet++}, MegaDepth \cite{li2018megadepth}, BlendedMVS \cite{yao2020blendedmvs}, TartanAir \cite{wangtartanair}, MVS-Synth \cite{huang2018deepmvs}, and ParallelDomain-4D \cite{van2024generative}, spanning both real and synthetic, indoor and outdoor scenes.
For \textbf{SL+}, we additionally add two fisheye datasets for training: ASE \cite{engel2023project} and KITTI360 \cite{liao2022kitti}.
For testing, we evaluate the performance on 3 fisheye datasets: ScanNet++ (fisheye) \cite{yeshwanth2023scannet++}, ADT \cite{pan2023aria}, and KITTI360 \cite{liao2022kitti}.

\paragraphtitle{Metrics.}
We evaluate the models on 4 different tasks and 15 quantitative metrics following \cite{wang2025pi,keetha2025mapanything,tirado2025anycalib}.
For camera pose estimation, we use Relative Rotation Accuracy (RRA), Relative Translation Accuracy (RTA), and Area Under Curve (AUC) of the accuracy-threshold curve at 30$^\circ$ as angular accuracy metrics and Absolute Trajectory Error (ATE), Relative Pose Error for translation (RPEt), and Relative Pose Error for rotation (RPEr) as distance error metrics.
For depth map estimation, we use Absolute Relative Error (Rel), Root Mean Squared Error (RMSE), and $\delta_1$ accuracy.
For point map estimation, we use Accuracy (Acc), Completeness (Comp), and Chamfer Distance (CD).
For FoV map estimation, we use median error for horizontal ($\text{hErr}$) and vertical ($\text{vErr}$) FoVs in degrees, as well as the Area Under Curve ($\text{AUC}$) up to a $10^\circ$ threshold for FoV.

\paragraphtitle{Models.}
We apply our method on three foundation reconstruction models, VGGT \cite{wang2025vggt}, $\pi^3$ \cite{wang2025pi}, and MapAnything \cite{keetha2025mapanything}, based on their public checkpoints.

\subsection{Results}

\paragraphtitle{Quantitative improvements on fisheye datasets.}
\Cref{tab:quantitative} summarizes the quantitative results on three datasets of fisheye sequences.
Fisheye3R yields consistent and substantial performance gains over the baseline models, demonstrating that calibration tokens can effectively bridge the geometric domain gap between perspective and fisheye imagery.
Remarkably, even self-supervised learning using only unlabeled perspective RGB images boosts performance on fisheye data by a large margin, achieving a 50\%+ improvement in 26/135 tests across all models and metrics.
This highlights our framework's potential to scale with larger corpora of unlabeled data for further enhancement.
The introduction of annotated perspective data further strengthens these results, with 37/135 metrics showing improvements exceeding 50\%.
Performance ultimately peaks when incorporating annotated fisheye data, where a 50\%+ improvement is achieved in 77/135 cases.
This shows that while Fisheye3R is highly effective in data-constrained regimes, it scales efficiently as more supervision becomes available.

\paragraphtitle{Qualitative comparisons.}
We qualitatively compare our method with previous unified feed-forward foundation models in \cref{fig:qualitative}. 
Our method achieves significantly more accurate 3D scene reconstruction from the sequence of fisheye images thanks to the calibration tokens.
While traditional foundation models are incapable of accurately predicting 3D scene geometry from distorted fisheye lenses, our method produces reconstructions with higher fidelity and better alignment with ground truth.

\begin{figure*}[t]
    \centering
    \includegraphics[width=\linewidth]{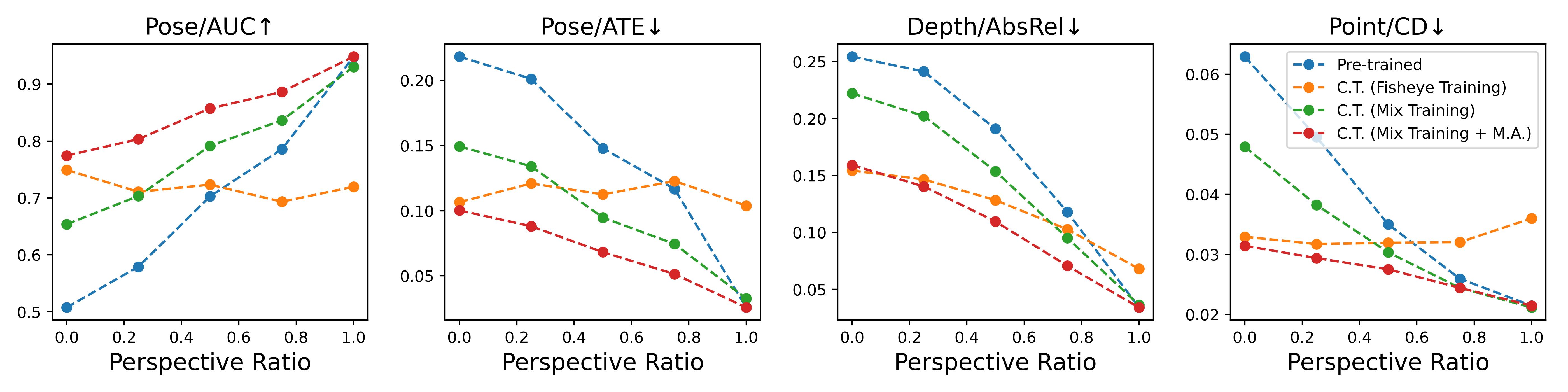}
    \vspace{-5mm}
    \caption{
        Performance analysis on hybrid sequences of ScanNet++ with varying perspective image ratios.
        \textbf{Pre-trained}: the vanilla MapAnything backbone without adaptation.
        \textbf{C.T. (Fisheye Training)}: calibration tokens trained on fully fisheye sequences.
        \textbf{C.T. (Mix Training)}: tokens trained on mixture sequences of both camera types.
        \textbf{C.T. (Mix Training + M.A.)}: masked attention introduced during mixed training.
    }
    \label{fig:mixture}
\end{figure*}

\begin{figure*}[t]
    \centering
    \includegraphics[width=\linewidth]{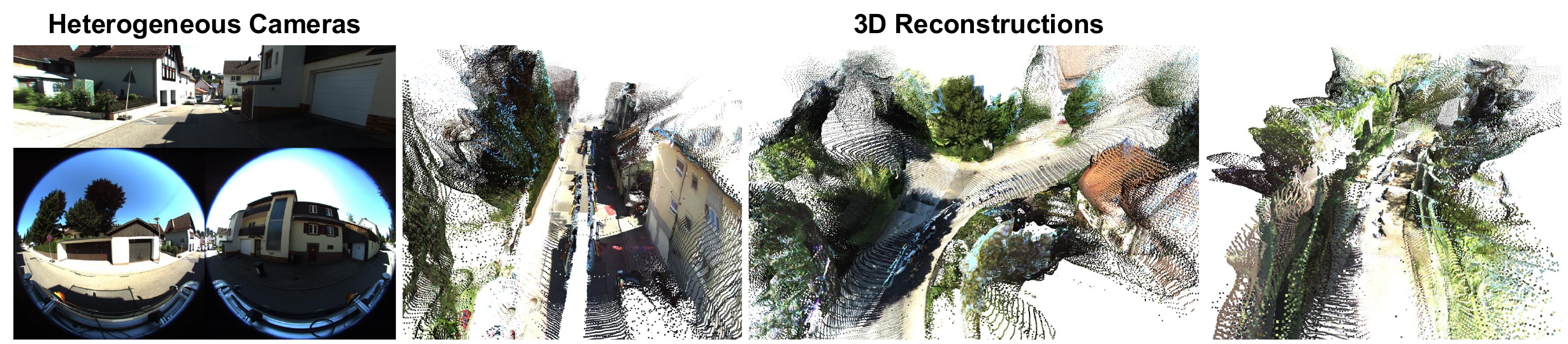}
    \vspace{-5mm}
    \caption{
        Reconstructions over a hybrid camera system in driving scenes \cite{liao2022kitti}.
        The left and right fisheye cameras capture the two sides of the scene, while the front-facing perspective camera bridges the two views.
        Without a model capable of handling heterogeneous cameras, reconstructions from the two fisheye views would remain disconnected.
    }
    \label{fig:mixture_kitti360}
\end{figure*}

\paragraphtitle{Performance on hybrid inputs.}
We analyze the calibration performance across hybrid sequences containing both perspective and fisheye frames in \cref{fig:mixture} and \cref{fig:mixture_kitti360}.
By varying the ratio of perspective images, \cref{fig:mixture} shows that while calibration tokens excel at optimizing pure fisheye sequences, they suffer from parameter drift caused by specialized training on fisheye images, leading to performance degradation on perspective images.
Conversely, while training tokens on mixed-camera sequences preserves perspective accuracy, it weakens adaptation to fisheye data.
By contrast, our proposed masked attention mechanism not only enables backwards compatibility by retaining the model's native performance on perspective data, but also unleashes the representational power of calibration tokens for fisheye adaptation, achieving the most robust performance across all perspective ratios.
The reconstructions in \cref{fig:mixture_kitti360} illustrate a practical deployment scenario of our mixed-camera model for autonomous driving, where a vehicle is equipped with heterogeneous cameras.
Leveraging the forward-facing perspective camera to connect the left and right fisheye views, our method reconstructs a single, geometrically consistent panoramic scene from hybrid sensors.

\begin{figure*}[t]
    \centering
    \includegraphics[width=\linewidth]{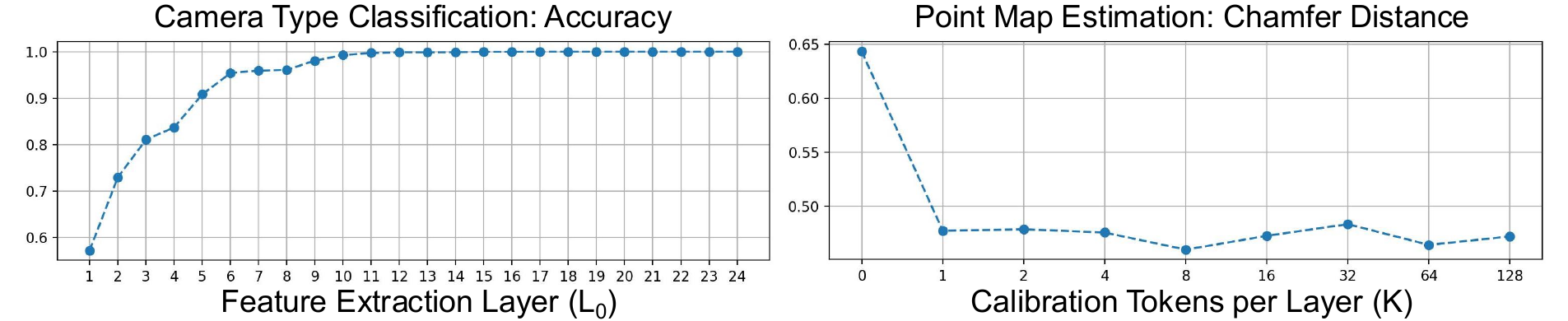}
    \vspace{-5mm}
    \caption{
        Ablation studies of hyperparameters.
        \textbf{Feature extraction depth} (left): Classification accuracy of camera types as a function of the number of frozen initial layers ($L_0$).
        \textbf{Token density} (right): Impact of the number of calibration tokens per layer ($K$) on reconstruction accuracy, measured by Chamfer Distance.
        Experiments are conducted on ScanNet++ with MapAnything backbone.
    }
    \label{fig:ablation}
\end{figure*}

\paragraphtitle{Ablation experiments.}
We present two ablation studies in \cref{fig:ablation}.
First, we analyze the feature extraction depth ($L_0$) required for robust camera type classification.
While deeper layers provide the class token with higher-level semantic information, increasing $L_0$ conversely reduces the number of subsequent layers ($L_1 - L_0$) available for feature alignment via calibration tokens.
As shown in \cref{fig:ablation} (left), classification accuracy saturates at approximately $L_0 = 12$ layers.
We identify this as the optimal bottleneck, as it reserves the remaining 12 encoder layers for calibration, thereby maximizing the potential for geometric adaptation.
Second, we evaluate the number of calibration tokens ($K$) per layer in \cref{fig:ablation} (right), which reveals that even a single token ($K=1$) per layer brings the majority of the performance gain, demonstrating the efficiency of the proposed token-based alignment.
While increasing the token count yields further marginal improvements, the reconstruction quality peaks approximately at $K = 8$.
Therefore, we adopt $L_0 = 12$ and $K = 8$ as the default configuration for the experiments.

\begin{figure*}[t]
    \centering
    \includegraphics[width=\linewidth]{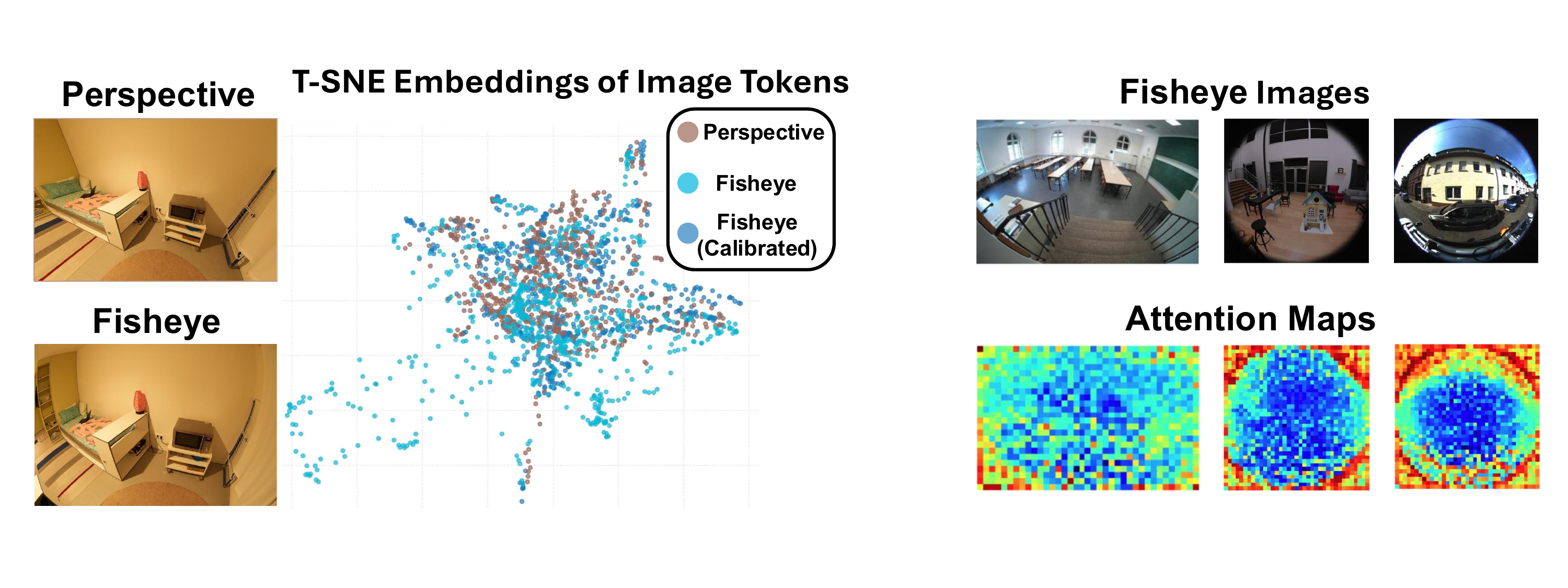}
    \vspace{-10mm}
    \caption{
        Visualizing the impact of calibration tokens on the feature space with MapAnything.
        \textbf{Feature alignment} (left): Image tokens from the last alternating attention layer of a perspective and fisheye image pair, with fisheye tokens derived from the model before and after calibration.
        \textbf{Attention maps} (right): Three fisheye test images and their average attention weights from image to calibration tokens in the image encoder.
    }
    \label{fig:features}
\end{figure*}

\paragraphtitle{Feature alignment.}
We employ t-SNE \cite{van2008visualizing} to visualize the patch-level embeddings of a perspective and fisheye image pair in \cref{fig:features} (left).
As shown in the embedding space, the original fisheye features (light blue) exhibit a distinct distributional shift and deviate significantly from the perspective manifold (brown).
This geometric domain gap explains the performance degradation observed when applying the pre-trained foundation model directly to fisheye data.
By introducing calibration tokens, the adapted fisheye embeddings (dark blue) are effectively realigned to the perspective distribution.
This demonstrates that the tokens are able to bridge the geometric gap by projecting fisheye patches into a shared representation space consistent with the model's original training distribution.

\paragraphtitle{Attention map visualization.}
To further understand the spatial influence of our framework, we visualize the attention weights that image tokens pay to the calibration tokens in \cref{fig:features} (right).
These attention maps serve as an indicator of the corrective influence exerted by the calibration tokens on specific image patches.
It can be observed that patches in the peripheral regions that suffer from the most significant radial distortion pay more attention to calibration tokens (red).
This suggests that the calibration tokens selectively target regions where the features are most divergent from the perspective distribution, prioritizing the correction of extreme geometric warping at fisheye image boundaries to restore global feature consistency.

\begin{table*}[t]
    \centering
    \resizebox{\columnwidth}{!}{
        \newcolumntype{M}{>{\centering\arraybackslash}m{20mm}}
        \begin{tabular}{c | MMM | MMMM}
            \toprule[1pt]
            Model
            & KB (OOD)
            & Fisheye624
            & MEI
            & Equidistant
            & Stereographic
            & Equiangular
            & Orthographic
            \\ \midrule
            $\pi^3$ CD$\downarrow$
            & 0.230 & 0.228 & 0.264 & 0.219 & 0.211 & 0.241 & 0.375 
            \\
            +Ours
            & 0.116 & 0.117 & 0.095 & 0.107 & 0.088 & 0.116 & 0.182 
            \\ \midrule
            Improvement
            & 49.7\% & 48.5\% & 64.1\% & 51.1\% & 58.1\% & 52.0\% & 51.5\%
            \\
            \bottomrule[1pt]
        \end{tabular}
    }
    \vspace{1mm}
    \caption{
        Reconstruction CD of $\pi^3$ on Stanford2D3DS with various projection models.
        Although our method employs Kannala-Brandt (KB) for synthesis, it generalizes well to KB with out-of-distribution (OOD) parameters and other projection models.
    }
    \label{tab:distortion_models}
\end{table*}

\paragraphtitle{Generalization to other projection models.}
To evaluate the projection model domain gap in a controlled setting and to eliminate dataset-dependent confounding effects, we render fisheye images from Stanford2D3DS dataset \cite{armeni2017joint} with varying projection models and measure the improvement of reconstruction quality of our approach in \cref{tab:distortion_models}.
The results indicate that although only Kannala-Brandt model is employed for fisheye synthesis, the learned calibration tokens generalize well to other projection models.

\begin{figure*}[t]
    \centering
    \includegraphics[width=\linewidth]{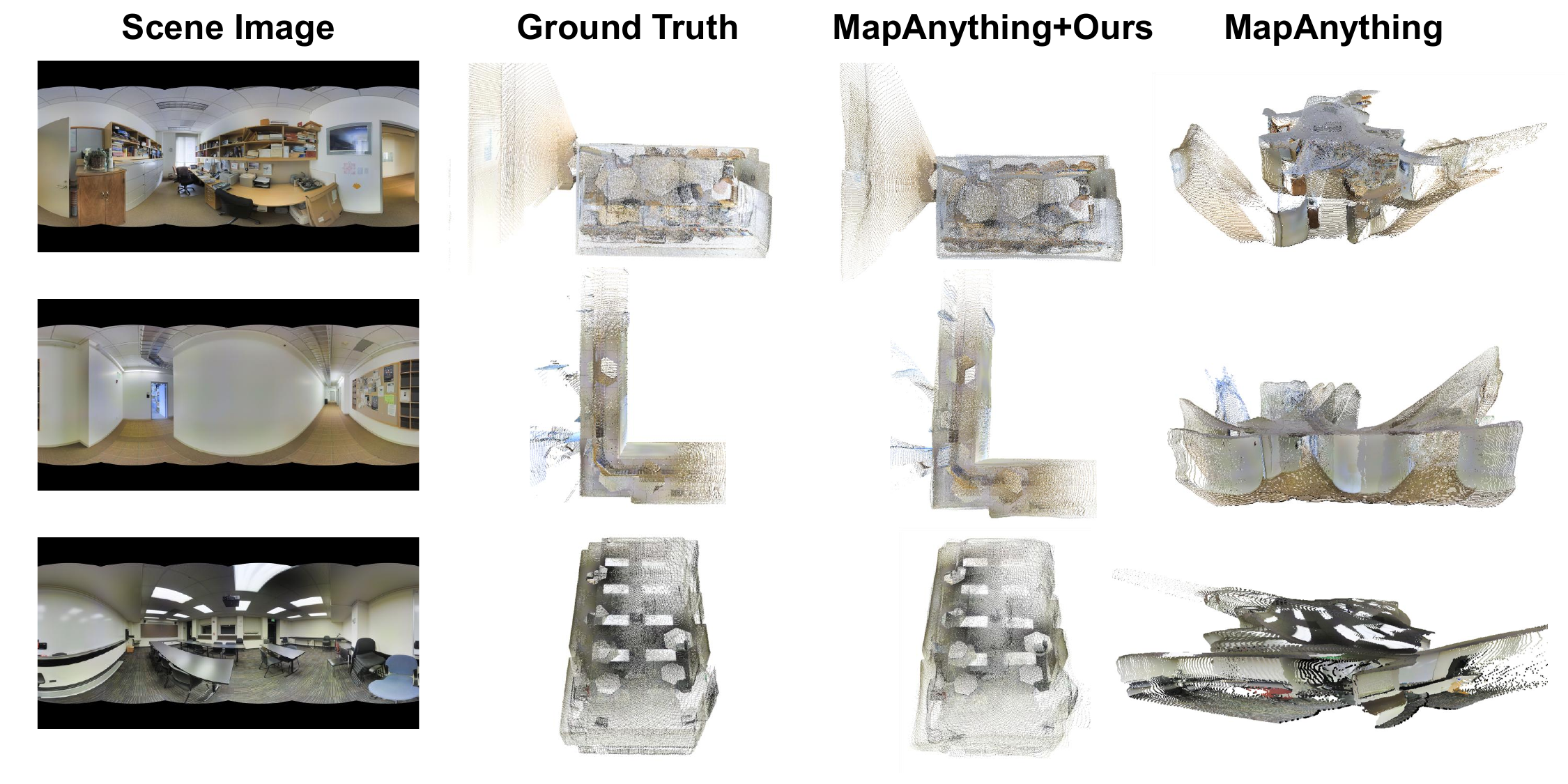}
    \vspace{-5mm}
    \caption{
        Generalization of calibration tokens to panoramic images.
    }
    \label{fig:vis_stanford2d3ds}
    \vspace{-2mm}
\end{figure*}

\paragraphtitle{Generalization to panoramic images.}
While this work primarily focuses on extending foundation models to fisheye imagery, calibration tokens have the potential to generalize to images from other camera models as well, such as panoramic images.
\Cref{fig:vis_stanford2d3ds} presents multi-view reconstructions from 360$^\circ$ images, where calibration tokens are trained on Matterport3D \cite{chang2017matterport3d} and tested on Stanford2D3DS \cite{armeni2017joint}.
The result implies that calibration tokens can be generalized to camera models beyond fisheye.

\section{Discussion}

Projective geometry describes the mapping between 3D points and 2D image pixels. Different camera lenses induce different projection functions, resulting in different spatial arrangements of pixels in the images formed. As existing models are trained on perspective images, they take on such biases implicitly. When tested on images with curvilinear distortion, even for 3D scenes previously observed, local operations such as patch embedding operate on a different set of visual patterns that induce a covariate shift in the latent features, leading to out-of-distribution behavior and performance degradation. As the change occurs in the value of the features distributed over the spatial domain, our work aims to recalibrate them with learnable tokens to align with the original training distribution of perspective features. While this can be viewed as a form of camera domain adaptation, our masked attention mechanism enables a single network to support different camera models by selectively controlling the activation of adaptation based on camera types.

\paragraphtitle{Limitations.}
The proposed method is limited to transformer architectures due to its dependency on token-based adaptation. The fidelity of self-supervised adaptation also depends on the original model's performance. While this is partially mitigated by the use of ground truth in the supervised setting, one may be limited for tasks that do not have annotations readily available.

\section*{Acknowledgments}
This work is supported by Google Gift Funding and NSF-2112562 Athena AI Institute.
Additionally, we thank Federico Tombari for his valuable guidance and constructive feedback during the course of this project.

\bibliographystyle{splncs04}
\bibliography{main,visionlab}

@String(ECCV  = {Eur. Conf. Comput. Vis.})

@String(ICLR  = {Int. Conf. Learn. Represent.})

@String(ECCV  = {ECCV})

@String(ICLR  = {ICLR})

@inproceedings{schonberger2016structure,
  title={Structure-from-motion revisited},
  author={Schonberger, Johannes L and Frahm, Jan-Michael},
  booktitle={Proceedings of the IEEE conference on computer vision and pattern recognition},
  pages={4104--4113},
  year={2016}
}

@inproceedings{pan2024global,
  title={Global structure-from-motion revisited},
  author={Pan, Linfei and Bar{\'a}th, D{\'a}niel and Pollefeys, Marc and Sch{\"o}nberger, Johannes L},
  booktitle={European Conference on Computer Vision},
  pages={58--77},
  year={2024},
  organization={Springer}
}

@inproceedings{gherardi2010improving,
  title={Improving the efficiency of hierarchical structure-and-motion},
  author={Gherardi, Riccardo and Farenzena, Michela and Fusiello, Andrea},
  booktitle={2010 IEEE computer society conference on computer vision and pattern recognition},
  pages={1594--1600},
  year={2010},
  organization={IEEE}
}

@inproceedings{wu2013towards,
  title={Towards linear-time incremental structure from motion},
  author={Wu, Changchang},
  booktitle={2013 International Conference on 3D Vision-3DV 2013},
  pages={127--134},
  year={2013},
  organization={IEEE}
}

@inproceedings{sweeney2015optimizing,
  title={Optimizing the viewing graph for structure-from-motion},
  author={Sweeney, Chris and Sattler, Torsten and Hollerer, Tobias and Turk, Matthew and Pollefeys, Marc},
  booktitle={Proceedings of the IEEE international conference on computer vision},
  pages={801--809},
  year={2015}
}

@article{agarwal2011building,
  title={Building rome in a day},
  author={Agarwal, Sameer and Furukawa, Yasutaka and Snavely, Noah and Simon, Ian and Curless, Brian and Seitz, Steven M and Szeliski, Richard},
  journal={Communications of the ACM},
  volume={54},
  number={10},
  pages={105--112},
  year={2011},
  publisher={ACM New York, NY, USA}
}

@inproceedings{frahm2010building,
  title={Building rome on a cloudless day},
  author={Frahm, Jan-Michael and Fite-Georgel, Pierre and Gallup, David and Johnson, Tim and Raguram, Rahul and Wu, Changchang and Jen, Yi-Hung and Dunn, Enrique and Clipp, Brian and Lazebnik, Svetlana and others},
  booktitle={European conference on computer vision},
  pages={368--381},
  year={2010},
  organization={Springer}
}

@incollection{snavely2006photo,
  title={Photo tourism: exploring photo collections in 3D},
  author={Snavely, Noah and Seitz, Steven M and Szeliski, Richard},
  booktitle={ACM siggraph 2006 papers},
  pages={835--846},
  year={2006}
}

@inproceedings{schonberger2016pixelwise,
  title={Pixelwise view selection for unstructured multi-view stereo},
  author={Sch{\"o}nberger, Johannes L and Zheng, Enliang and Frahm, Jan-Michael and Pollefeys, Marc},
  booktitle={European conference on computer vision},
  pages={501--518},
  year={2016},
  organization={Springer}
}

@inproceedings{yao2018mvsnet,
  title={Mvsnet: Depth inference for unstructured multi-view stereo},
  author={Yao, Yao and Luo, Zixin and Li, Shiwei and Fang, Tian and Quan, Long},
  booktitle={Proceedings of the European conference on computer vision (ECCV)},
  pages={767--783},
  year={2018}
}

@inproceedings{gu2020cascade,
  title={Cascade cost volume for high-resolution multi-view stereo and stereo matching},
  author={Gu, Xiaodong and Fan, Zhiwen and Zhu, Siyu and Dai, Zuozhuo and Tan, Feitong and Tan, Ping},
  booktitle={Proceedings of the IEEE/CVF conference on computer vision and pattern recognition},
  pages={2495--2504},
  year={2020}
}

@inproceedings{ding2022transmvsnet,
  title={Transmvsnet: Global context-aware multi-view stereo network with transformers},
  author={Ding, Yikang and Yuan, Wentao and Zhu, Qingtian and Zhang, Haotian and Liu, Xiangyue and Wang, Yuanjiang and Liu, Xiao},
  booktitle={Proceedings of the IEEE/CVF conference on computer vision and pattern recognition},
  pages={8585--8594},
  year={2022}
}

@inproceedings{wang2021patchmatchnet,
  title={Patchmatchnet: Learned multi-view patchmatch stereo},
  author={Wang, Fangjinhua and Galliani, Silvano and Vogel, Christoph and Speciale, Pablo and Pollefeys, Marc},
  booktitle={Proceedings of the IEEE/CVF conference on computer vision and pattern recognition},
  pages={14194--14203},
  year={2021}
}

@inproceedings{zhao2023mvpsnet,
  title={Mvpsnet: Fast generalizable multi-view photometric stereo},
  author={Zhao, Dongxu and Lichy, Daniel and Perrin, Pierre-Nicolas and Frahm, Jan-Michael and Sengupta, Soumyadip},
  booktitle={Proceedings of the IEEE/CVF International Conference on Computer Vision},
  pages={12525--12536},
  year={2023}
}

@inproceedings{wang2024dust3r,
  title={Dust3r: Geometric 3d vision made easy},
  author={Wang, Shuzhe and Leroy, Vincent and Cabon, Yohann and Chidlovskii, Boris and Revaud, Jerome},
  booktitle={Proceedings of the IEEE/CVF conference on computer vision and pattern recognition},
  pages={20697--20709},
  year={2024}
}

@inproceedings{leroy2024grounding,
  title={Grounding image matching in 3d with mast3r},
  author={Leroy, Vincent and Cabon, Yohann and Revaud, J{\'e}r{\^o}me},
  booktitle={European conference on computer vision},
  pages={71--91},
  year={2024},
  organization={Springer}
}

@article{zhang2024monst3r,
  title={Monst3r: A simple approach for estimating geometry in the presence of motion},
  author={Zhang, Junyi and Herrmann, Charles and Hur, Junhwa and Jampani, Varun and Darrell, Trevor and Cole, Forrester and Sun, Deqing and Yang, Ming-Hsuan},
  journal={arXiv preprint arXiv:2410.03825},
  year={2024}
}

@inproceedings{yang2025fast3r,
  title={Fast3r: Towards 3d reconstruction of 1000+ images in one forward pass},
  author={Yang, Jianing and Sax, Alexander and Liang, Kevin J and Henaff, Mikael and Tang, Hao and Cao, Ang and Chai, Joyce and Meier, Franziska and Feiszli, Matt},
  booktitle={Proceedings of the Computer Vision and Pattern Recognition Conference},
  pages={21924--21935},
  year={2025}
}

@inproceedings{wang2025continuous,
  title={Continuous 3d perception model with persistent state},
  author={Wang, Qianqian and Zhang, Yifei and Holynski, Aleksander and Efros, Alexei A and Kanazawa, Angjoo},
  booktitle={Proceedings of the Computer Vision and Pattern Recognition Conference},
  pages={10510--10522},
  year={2025}
}

@inproceedings{wang2025vggt,
  title={Vggt: Visual geometry grounded transformer},
  author={Wang, Jianyuan and Chen, Minghao and Karaev, Nikita and Vedaldi, Andrea and Rupprecht, Christian and Novotny, David},
  booktitle={Proceedings of the Computer Vision and Pattern Recognition Conference},
  pages={5294--5306},
  year={2025}
}

@article{wang2025pi,
  title={$\pi^3$: Permutation-Equivariant Visual Geometry Learning},
  author={Wang, Yifan and Zhou, Jianjun and Zhu, Haoyi and Chang, Wenzheng and Zhou, Yang and Li, Zizun and Chen, Junyi and Pang, Jiangmiao and Shen, Chunhua and He, Tong},
  journal={arXiv preprint arXiv:2507.13347},
  year={2025}
}

@article{keetha2025mapanything,
  title={Mapanything: Universal feed-forward metric 3d reconstruction},
  author={Keetha, Nikhil and M{\"u}ller, Norman and Sch{\"o}nberger, Johannes and Porzi, Lorenzo and Zhang, Yuchen and Fischer, Tobias and Knapitsch, Arno and Zauss, Duncan and Weber, Ethan and Antunes, Nelson and others},
  journal={arXiv preprint arXiv:2509.13414},
  year={2025}
}

@inproceedings{yeshwanth2023scannet++,
  title={Scannet++: A high-fidelity dataset of 3d indoor scenes},
  author={Yeshwanth, Chandan and Liu, Yueh-Cheng and Nie{\ss}ner, Matthias and Dai, Angela},
  booktitle={Proceedings of the IEEE/CVF International Conference on Computer Vision},
  pages={12--22},
  year={2023}
}

@inproceedings{pan2023aria,
  title={Aria digital twin: A new benchmark dataset for egocentric 3d machine perception},
  author={Pan, Xiaqing and Charron, Nicholas and Yang, Yongqian and Peters, Scott and Whelan, Thomas and Kong, Chen and Parkhi, Omkar and Newcombe, Richard and Ren, Yuheng Carl},
  booktitle={Proceedings of the IEEE/CVF International Conference on Computer Vision},
  pages={20133--20143},
  year={2023}
}

@article{engel2023project,
  title={Project aria: A new tool for egocentric multi-modal ai research},
  author={Engel, Jakob and Somasundaram, Kiran and Goesele, Michael and Sun, Albert and Gamino, Alexander and Turner, Andrew and Talattof, Arjang and Yuan, Arnie and Souti, Bilal and Meredith, Brighid and others},
  journal={arXiv preprint arXiv:2308.13561},
  year={2023}
}

@article{liao2022kitti,
  title={Kitti-360: A novel dataset and benchmarks for urban scene understanding in 2d and 3d},
  author={Liao, Yiyi and Xie, Jun and Geiger, Andreas},
  journal={IEEE Transactions on Pattern Analysis and Machine Intelligence},
  volume={45},
  number={3},
  pages={3292--3310},
  year={2022},
  publisher={IEEE}
}

@inproceedings{li2018megadepth,
  title={Megadepth: Learning single-view depth prediction from internet photos},
  author={Li, Zhengqi and Snavely, Noah},
  booktitle={Proceedings of the IEEE conference on computer vision and pattern recognition},
  pages={2041--2050},
  year={2018}
}

@inproceedings{yao2020blendedmvs,
  title={Blendedmvs: A large-scale dataset for generalized multi-view stereo networks},
  author={Yao, Yao and Luo, Zixin and Li, Shiwei and Zhang, Jingyang and Ren, Yufan and Zhou, Lei and Fang, Tian and Quan, Long},
  booktitle={Proceedings of the IEEE/CVF conference on computer vision and pattern recognition},
  pages={1790--1799},
  year={2020}
}

@inproceedings{wangtartanair,
  title={Tartanair: A dataset to push the limits of visual slam. In 2020 IEEE},
  author={Wang, Wenshan and Zhu, Delong and Wang, Xiangwei and Hu, Yaoyu and Qiu, Yuheng and Wang, Chen and Hu, Yafei and Kapoor, Ashish and Scherer, Sebastian},
  booktitle={RSJ International Conference on Intelligent Robots and Systems (IROS)},
  pages={4909--4916}
}

@inproceedings{huang2018deepmvs,
  title={Deepmvs: Learning multi-view stereopsis},
  author={Huang, Po-Han and Matzen, Kevin and Kopf, Johannes and Ahuja, Narendra and Huang, Jia-Bin},
  booktitle={Proceedings of the IEEE conference on computer vision and pattern recognition},
  pages={2821--2830},
  year={2018}
}

@inproceedings{van2024generative,
  title={Generative camera dolly: Extreme monocular dynamic novel view synthesis},
  author={Van Hoorick, Basile and Wu, Rundi and Ozguroglu, Ege and Sargent, Kyle and Liu, Ruoshi and Tokmakov, Pavel and Dave, Achal and Zheng, Changxi and Vondrick, Carl},
  booktitle={European Conference on Computer Vision},
  pages={313--331},
  year={2024},
  organization={Springer}
}

@inproceedings{caruso2015large,
  title={Large-scale direct SLAM for omnidirectional cameras},
  author={Caruso, David and Engel, Jakob and Cremers, Daniel},
  booktitle={2015 IEEE/RSJ international conference on intelligent robots and systems (IROS)},
  pages={141--148},
  year={2015},
  organization={IEEE}
}

@article{matsuki2018omnidirectional,
  title={Omnidirectional DSO: Direct sparse odometry with fisheye cameras},
  author={Matsuki, Hidenobu and Von Stumberg, Lukas and Usenko, Vladyslav and St{\"u}ckler, J{\"o}rg and Cremers, Daniel},
  journal={IEEE Robotics and Automation Letters},
  volume={3},
  number={4},
  pages={3693--3700},
  year={2018},
  publisher={IEEE}
}

@inproceedings{scaramuzza2006flexible,
  title={A flexible technique for accurate omnidirectional camera calibration and structure from motion},
  author={Scaramuzza, Davide and Martinelli, Agostino and Siegwart, Roland},
  booktitle={Fourth IEEE International Conference on Computer Vision Systems (ICVS'06)},
  pages={45--45},
  year={2006},
  organization={IEEE}
}

@inproceedings{mei2007single,
  title={Single view point omnidirectional camera calibration from planar grids},
  author={Mei, Christopher and Rives, Patrick},
  booktitle={Proceedings 2007 IEEE International Conference on Robotics and Automation},
  pages={3945--3950},
  year={2007},
  organization={IEEE}
}

@inproceedings{geyer2000unifying,
  title={A unifying theory for central panoramic systems and practical implications},
  author={Geyer, Christopher and Daniilidis, Kostas},
  booktitle={European conference on computer vision},
  pages={445--461},
  year={2000},
  organization={Springer}
}

@article{kannala2006generic,
  title={A generic camera model and calibration method for conventional, wide-angle, and fish-eye lenses},
  author={Kannala, Juho and Brandt, Sami S},
  journal={IEEE transactions on pattern analysis and machine intelligence},
  volume={28},
  number={8},
  pages={1335--1340},
  year={2006},
  publisher={IEEE}
}

@inproceedings{zhao2025fisheyedepth,
  title={Fisheyedepth: A real scale self-supervised depth estimation model for fisheye camera},
  author={Zhao, Guoyang and Liu, Yuxuan and Qi, Weiqing and Ma, Fulong and Liu, Ming and Ma, Jun},
  booktitle={2025 IEEE International Conference on Robotics and Automation (ICRA)},
  pages={3780--3787},
  year={2025},
  organization={IEEE}
}

@inproceedings{guo2025depth,
  title={Depth any camera: Zero-shot metric depth estimation from any camera},
  author={Guo, Yuliang and Garg, Sparsh and Miangoleh, S Mahdi H and Huang, Xinyu and Ren, Liu},
  booktitle={Proceedings of the Computer Vision and Pattern Recognition Conference},
  pages={26996--27006},
  year={2025}
}

@article{li2021omnidirectional,
  title={Omnidirectional stereo depth estimation based on spherical deep network},
  author={Li, Ming and Hu, Xuejiao and Dai, Jingzhao and Li, Yang and Du, Sidan},
  journal={Image and Vision Computing},
  volume={114},
  pages={104264},
  year={2021},
  publisher={Elsevier}
}

@inproceedings{wang2020360sd,
  title={360sd-net: 360 stereo depth estimation with learnable cost volume},
  author={Wang, Ning-Hsu and Solarte, Bolivar and Tsai, Yi-Hsuan and Chiu, Wei-Chen and Sun, Min},
  booktitle={2020 IEEE International Conference on Robotics and Automation (ICRA)},
  pages={582--588},
  year={2020},
  organization={IEEE}
}

@article{zhao2025fastvidar,
  title={FastViDAR: Real-Time Omnidirectional Depth Estimation via Alternative Hierarchical Attention},
  author={Zhao, Hangtian and Chen, Xiang and Li, Yizhe and Wang, Qianhao and Lu, Haibo and Gao, Fei},
  journal={arXiv preprint arXiv:2509.23733},
  year={2025}
}

@inproceedings{deng2025omnistereo,
  title={OmniStereo: Real-time Omnidireactional Depth Estimation with Multiview Fisheye Cameras},
  author={Deng, Jiaxi and Wang, Yushen and Meng, Haitao and Hou, Zuoxun and Chang, Yi and Chen, Gang},
  booktitle={Proceedings of the Computer Vision and Pattern Recognition Conference},
  pages={1003--1012},
  year={2025}
}

@inproceedings{veicht2024geocalib,
  title={Geocalib: Learning single-image calibration with geometric optimization},
  author={Veicht, Alexander and Sarlin, Paul-Edouard and Lindenberger, Philipp and Pollefeys, Marc},
  booktitle={European Conference on Computer Vision},
  pages={1--20},
  year={2024},
  organization={Springer}
}

@inproceedings{tirado2025anycalib,
  title={AnyCalib: On-manifold learning for model-agnostic single-view camera calibration},
  author={Tirado-Gar{\'\i}n, Javier and Civera, Javier},
  booktitle={Proceedings of the IEEE/CVF International Conference on Computer Vision},
  pages={8044--8055},
  year={2025}
}

@inproceedings{lichy2024fova,
  title={Fova-depth: Field-of-view agnostic depth estimation for cross-dataset generalization},
  author={Lichy, Daniel and Su, Hang and Badki, Abhishek and Kautz, Jan and Gallo, Orazio},
  booktitle={2024 International Conference on 3D Vision (3DV)},
  pages={1--10},
  year={2024},
  organization={IEEE}
}

@inproceedings{piccinelli2025unik3d,
  title={UniK3D: Universal Camera Monocular 3D Estimation},
  author={Piccinelli, Luigi and Sakaridis, Christos and Segu, Mattia and Yang, Yung-Hsu and Li, Siyuan and Abbeloos, Wim and Van Gool, Luc},
  booktitle={Proceedings of the Computer Vision and Pattern Recognition Conference},
  pages={1028--1039},
  year={2025}
}

@article{dosovitskiy2020image,
  title={An image is worth 16x16 words: Transformers for image recognition at scale},
  author={Dosovitskiy, Alexey and Beyer, Lucas and Kolesnikov, Alexander and Weissenborn, Dirk and Zhai, Xiaohua and Unterthiner, Thomas and Dehghani, Mostafa and Minderer, Matthias and Heigold, Georg and Gelly, Sylvain and others},
  journal={arXiv preprint arXiv:2010.11929},
  year={2020}
}

@article{oquab2023dinov2,
  title={Dinov2: Learning robust visual features without supervision},
  author={Oquab, Maxime and Darcet, Timoth{\'e}e and Moutakanni, Th{\'e}o and Vo, Huy and Szafraniec, Marc and Khalidov, Vasil and Fernandez, Pierre and Haziza, Daniel and Massa, Francisco and El-Nouby, Alaaeldin and others},
  journal={arXiv preprint arXiv:2304.07193},
  year={2023}
}

@article{van2008visualizing,
  title={Visualizing data using t-SNE.},
  author={Van der Maaten, Laurens and Hinton, Geoffrey},
  journal={Journal of machine learning research},
  volume={9},
  number={11},
  year={2008}
}

@inproceedings{schops2019bad,
  title={Bad slam: Bundle adjusted direct rgb-d slam},
  author={Schops, Thomas and Sattler, Torsten and Pollefeys, Marc},
  booktitle={Proceedings of the IEEE/CVF Conference on Computer Vision and Pattern Recognition},
  pages={134--144},
  year={2019}
}

@article{hu2022lora,
  title={Lora: Low-rank adaptation of large language models.},
  author={Hu, Edward J and Shen, Yelong and Wallis, Phillip and Allen-Zhu, Zeyuan and Li, Yuanzhi and Wang, Shean and Wang, Liang and Chen, Weizhu and others},
  journal={Iclr},
  volume={1},
  number={2},
  pages={3},
  year={2022}
}

@article{chang2017matterport3d,
  title={Matterport3d: Learning from rgb-d data in indoor environments},
  author={Chang, Angel and Dai, Angela and Funkhouser, Thomas and Halber, Maciej and Niessner, Matthias and Savva, Manolis and Song, Shuran and Zeng, Andy and Zhang, Yinda},
  journal={arXiv preprint arXiv:1709.06158},
  year={2017}
}

@article{armeni2017joint,
  title={Joint 2d-3d-semantic data for indoor scene understanding},
  author={Armeni, Iro and Sax, Sasha and Zamir, Amir R and Savarese, Silvio},
  journal={arXiv preprint arXiv:1702.01105},
  year={2017}
}

@article{umeyama2002least,
  title={Least-squares estimation of transformation parameters between two point patterns},
  author={Umeyama, Shinji},
  journal={IEEE Transactions on pattern analysis and machine intelligence},
  volume={13},
  number={4},
  pages={376--380},
  year={2002},
  publisher={IEEE}
}

@inproceedings{besl1992method,
  title={Method for registration of 3-D shapes},
  author={Besl, Paul J and McKay, Neil D},
  booktitle={Sensor fusion IV: control paradigms and data structures},
  volume={1611},
  pages={586--606},
  year={1992},
  organization={Spie}
}

@inproceedings{gangopadhyay2025extending,
  title={Extending foundational monocular depth estimators to fisheye cameras with calibration tokens},
  author={Gangopadhyay, Suchisrit and Kim, Jung-Hee and Chen, Xien and Rim, Patrick and Park, Hyoungseob and Wong, Alex},
  booktitle={Proceedings of the IEEE/CVF International Conference on Computer Vision},
  year={2025}
}

@inproceedings{gangopadhyay2026from,
  title={From perspective to fisheye depth estimation and open-vocabulary segmentation},
  author={Gangopadhyay, Suchisrit and Wong, Alex},
  booktitle={European Conference on Computer Vision},
  year={2026},
  organization={Springer}
}

@inproceedings{wu2024augundo,
  title={Augundo: Scaling up augmentations for monocular depth completion and estimation},
  author={Wu, Yangchao and Liu, Tian Yu and Park, Hyoungseob and Soatto, Stefano and Lao, Dong and Wong, Alex},
  booktitle={European Conference on Computer Vision},
  pages={274--293},
  year={2024},
  organization={Springer}
}

\maketitlesupplementary

\section{Implementation Details}

\paragraphtitle{Datasets.}
We report the dataset details in \cref{tab:datasets}.
For training with perspective data, we adopt the training split of \cite{keetha2025mapanything} on these 6 datasets.
For training with fisheye data, we select 30 scenes from ASE and 6 scenes from KITTI360 for \textbf{SL+}.
For evaluation, we use the test split of \cite{keetha2025mapanything} for ScanNet++, all the scenes of ADT, and 3 scenes of KITTI360.
Train-test split is enforced at the scene level to ensure no test sequence is seen by the model during training.

\paragraphtitle{Sequence sampling.}
The sampling scheme roughly follows \cite{keetha2025mapanything}.
The sequence length is randomly sampled from 2--24 images per sequence during training, and from 4, 8, 16, and 32 during evaluation.
A pairwise covisibility matrix is pre-computed for each scene based on the ground-truth depth maps beforehand for all datasets except KITTI360 with extra-long sequences, i.e., from each source image, the percentage of its pixels visible from each target image is calculated.
Then, after the first reference frame is randomly selected from a scene, the remaining frames are added incrementally by including images with at least 25\% of pixels covisible from any sampled images.
For KITTI360, due to its large number of frames per scene, we randomly sample the remaining frames uniformly from an interval centered at the first frame, with the frames ordered by their timestamps.
During training, we set a maximum number of images per GPU to 24, thus equivalently 1--12 sequences per GPU, depending on the sampled sequence length.
With 4 GPUs for training, the model is trained with an effective batch of 4--48 sequences per iteration.

\paragraphtitle{Image processing.}
Training augmentations include random resizing, cropping, color jittering, gray-scale conversion, and Gaussian blur.
Images are resized such that the longer edge is 518 pixels, while the shorter edge is adjusted to the nearest multiple of 14 that preserves the aspect ratio.

\paragraphtitle{Distortion synthesis.}
We apply Kannala-Brandt model to synthesize fisheye distortions from perspective images.
Specifically, each 3D ray direction $\boldsymbol{d} = (x, y, z)$ is mapped to the image plane by first computing the incident angle
\begin{equation}
\theta = \arctan\frac{\sqrt{x^2 + y^2}}{z},
\end{equation}
and modeling the radial distance from the principal point as a polynomial of $\theta$:
\begin{equation}
r(\theta) = \theta + k_1 \theta^3 + k_2 \theta^5 + k_3 \theta^7 + k_4 \theta^9 ,
\end{equation}
where $k_1, k_2, k_3, k_4$ are distortion coefficients.
The final corresponding pixel coordinates $(u, v)$ in the fisheye image are given by
\begin{equation}
u = f_x \frac{x}{\sqrt{x^2+y^2}} r(\theta) + c_x, \qquad
v = f_y \frac{y}{\sqrt{x^2+y^2}} r(\theta) + c_y ,
\end{equation}
where $(f_x,f_y)$ and $(c_x,c_y)$ denote the focal lengths and the principal point.
During training, we randomly set the focal length to be [1, 1.2] times the original perspective focal length and shift the principal point by [-10, 10] pixels from the original principal point, with coefficients being uniformly sampled from $k_1, k_2, k_3 \in (-0.5, 0.5)$ and $k_4 \in (-0.05, 0.05)$ to synthesize different distortions.
For mixture training with both camera types, each sequence is randomly selected to be either fully perspective, fully fisheye, or hybrid.
For a hybrid sequence, each frame has a 0.5 probability of being distorted.

\begin{table}[t]
    \centering
    \small
    \begin{tabular}{cc|cccc}
    \toprule
        Split & Dataset & Type & Environment & Scenes & Images/Scene \\
        \midrule
        \multirow{6}{*}{\shortstack{Train\\(Perspective)}}
        & ScanNet++ \cite{yeshwanth2023scannet++} & Real & Indoor & 896 & 1059.6 \\
        & MegaDepth \cite{li2018megadepth} & Real & Outdoor & 266 & 153.5 \\
        & BlendedMVS \cite{yao2020blendedmvs} & Synthetic & Both & 450 & 232.5 \\
        & TartanAir \cite{wangtartanair} & Synthetic & Both & 46 & 7205.3 \\
        & MVS-Synth \cite{huang2018deepmvs} & Synthetic & Outdoor & 114 & 100.0 \\
        & ParallelDomain-4D \cite{van2024generative} & Synthetic & Outdoor & 1452 & 947.7 \\
        \midrule
        \multirow{2}{*}{\shortstack{Train\\(Fisheye)}}
        & ASE \cite{engel2023project} & Synthetic & Indoor & 30 & 608.2 \\
        & KITTI360 \cite{liao2022kitti} & Real & Outdoor & 6 & 16098.0 \\
        \midrule
        \multirow{3}{*}{\shortstack{Test\\(Fisheye)}}
        & ScanNet++ \cite{yeshwanth2023scannet++} & Real & Indoor & 30 & 1435.3 \\
        & ADT \cite{pan2023aria} & Real & Indoor & 236 & 70.3 \\
        & KITTI360 \cite{liao2022kitti} & Real & Outdoor & 3 & 15060.0 \\
    \bottomrule
    \end{tabular}
    \vspace{1mm}
    \caption{
        Summary of datasets used in our experiments. We report dataset type, environment, number of scenes, and the average number of images per scene.
    }
    \label{tab:datasets}
\end{table}

\paragraphtitle{Training.}
The 3D reconstruction loss function $\mathcal{L}$ used by the three learning schemes in \cref{eq:loss_ssl,eq:loss_sl,eq:loss_slp} follows the implementation of the MapAnything codebase \cite{keetha2025mapanything}, including supervision on camera poses, ray directions, depth maps, point maps, etc.
For calibration token implementation, we insert $K = 8$ tokens to each transformer encoder layer in the image encoder and alternating attention module, except the first $L_0 = 12$ layers of the image encoder, as more calibration tokens yield negligible performance gain.
Ablation in \cref{tab:ablation_components} shows that excluding the first 12 layers from calibration would not degrade the performance significantly.
For each model, we freeze its original backbone and initialize calibration tokens with values sampled from a normal distribution with mean zero and standard deviation 1e-6.
The calibration tokens are then trained for $40{,}000$ iterations by AdamW optimizer with a learning rate gradually decaying from 1e-5 to 1e-7.
Gradient checkpointing is enabled on the encoders to reduce the GPU memory consumption during training.

\paragraphtitle{Evaluation.}
We generally follow the protocols of prior work \cite{wang2025pi,keetha2025mapanything,tirado2025anycalib} for evaluating 3D reconstruction.
Specifically, predicted depth maps are aligned to ground truth by applying a scale and shift per sequence.
The predicted point maps are aligned to the ground truth using the Umeyama algorithm \cite{umeyama2002least} to estimate a similarity transformation, followed by a refinement step by Iterative Closest Point (ICP) \cite{besl1992method}.
Then, the point cloud metrics are computed based on the mean of nearest-neighbor distances between the predicted and ground-truth points.

\paragraphtitle{Camera type classification.}
The camera type classifier is implemented as a simple linear model (i.e., logistic regression), which is pre-trained prior to learning the calibration tokens and converges in one minute with no GPU, achieving 99.9\%+ accuracy on the test data.
This implies that distinguishing fisheye images from perspective is relatively straightforward from their features, due to their distinct distortion patterns.
A more complex classifier could potentially yield further improvements, but given that the classification accuracy has already saturated on our test data, we adopt a simple linear classifier for efficiency.

\paragraphtitle{Masked attention implementation.}
The masked attention mechanism is implemented by the \path{attn_mask} argument in \path{torch.nn.functional.scaled_dot_product_attention()} function in PyTorch.

\paragraphtitle{Hardware configurations.}
The models are trained on a machine with 4 NVIDIA RTX A6000 GPUs (48GiB each) and 56 dual Intel Xeon Gold 6258R CPU cores.

\begin{table}[t]
    \centering
    \resizebox{\textwidth}{!}{
        \newcolumntype{M}{>{\centering\arraybackslash}m{9.5mm}}
        \begin{tabular}{cccc | cc | *{3}{MM|} MM }
        \toprule
            \multicolumn{4}{c|}{\multirow{2}{*}{Method}}
            & Trainable & Memory
            & \multicolumn{2}{c|}{Pose}
            & \multicolumn{2}{c|}{Depth}
            & \multicolumn{2}{c|}{Point}
            & \multicolumn{2}{c}{FoV}
            \\
            \cline{7-14}
            & & & & Parameters & (GiB)
            & RPEt$\downarrow$ & RPEr$\downarrow$ 
            & Rel$\downarrow$ & $\delta_1\uparrow$ 
            & Acc$\downarrow$ & Comp$\downarrow$ 
            & hErr$\downarrow$ & vErr$\downarrow$ 
            \\
            \midrule
            \multicolumn{4}{c|}{Baseline} & 1,228,491,222 & >48
            & 0.310 & 8.851 & 0.274 & 0.672 & 0.091 & 0.076 & 6.631 & 4.821 \\
            \midrule
            \multirow{8}{*}{\rotatebox{90}{C.T. ($K = 8$)}}
            & E$^*$ &  &  & 294,912 & 37.12
            & 0.144 & 4.595 & 0.181 & 0.798 & 0.064 & 0.048 & 3.546 & 2.358 \\
            & E &  &  & 147,456 & 34.85
            & 0.147 & 4.680 & 0.180 & 0.802 & 0.063 & 0.049 & 3.548 & 2.392 \\
            & & F &  & 98,304 & 32.02
            & 0.143 & 4.658 & 0.177 & 0.808 & 0.060 & 0.047 & 3.412 & 2.226 \\
            & &  & G & 98,304 & 32.58
            & 0.148 & 4.760 & 0.179 & 0.808 & 0.060 & 0.047 & 3.419 & 2.204 \\
            & E & F &  & 245,760 & 34.86
            & 0.136 & 4.516 & 0.175 & 0.808 & 0.059 & 0.046 & 3.351 & 2.180 \\
            & E &  & G & 245,760 & 35.23
            & 0.140 & 4.556 & 0.174 & 0.809 & 0.059 & 0.046 & 3.289 & 2.073 \\
            & & F & G & 196,608 & 32.59
            & 0.138 & 4.581 & 0.176 & 0.806 & 0.058 & \textbf{0.045} & 3.295 & 2.098 \\
            & E & F & G & 344,064 & 35.24
            & \textbf{0.134} & \textbf{4.370} & 0.171 & \textbf{0.810} & 0.058 & \textbf{0.045} & \textbf{3.264} & \textbf{2.074} \\
            \midrule
            \multirow{4}{*}{\rotatebox{90}{LoRA}}
            & \multicolumn{3}{c|}{$r = 4$} & 1,032,192 & 34.85
            & 0.142 & 4.603 & 0.178 & 0.804 & 0.060 & 0.046 & 3.668 & 2.394 \\
            & \multicolumn{3}{c|}{$r = 8$} & 2,064,384 & 34.86
            & 0.142 & 4.560 & 0.173 & 0.808 & 0.059 & 0.046 & 3.454 & 2.242 \\
            & \multicolumn{3}{c|}{$r = 16$} & 4,128,768 & 34.88
            & 0.139 & 4.494 & 0.172 & 0.808 & 0.058 & 0.046 & 3.380 & 2.177 \\
            & \multicolumn{3}{c|}{$r = 32$} & 8,257,536 & 34.93
            & 0.137 & 4.410 & \textbf{0.170} & \textbf{0.810} & \textbf{0.057} & \textbf{0.045} & 3.398 & 2.198 \\
        \bottomrule
        \end{tabular}
    }
    \vspace{1mm}
    \caption{
        Ablation study of calibration tokens (C.T.) inserted to different modules and comparison to LoRA (with different ranks $r$).
        Reconstruction metrics are reported on MapAnything model and fully fisheye sequences of ScanNet++.
        Modules to calibrate: E$^*$ = image encoder (all $L_1 = 24$ layers); E = image encoder (the last $L_1 - L_0 = 12$ layers), F = frame-wise attention layers in alternating attention, G = global attention layers in alternating attention.
        $K = 8$ calibration tokens are inserted to each layer.
    }
    \label{tab:ablation_components}
\end{table}

\section{Further Discussions}

\begin{figure*}[ht!]
    \centering
    \includegraphics[width=\linewidth]{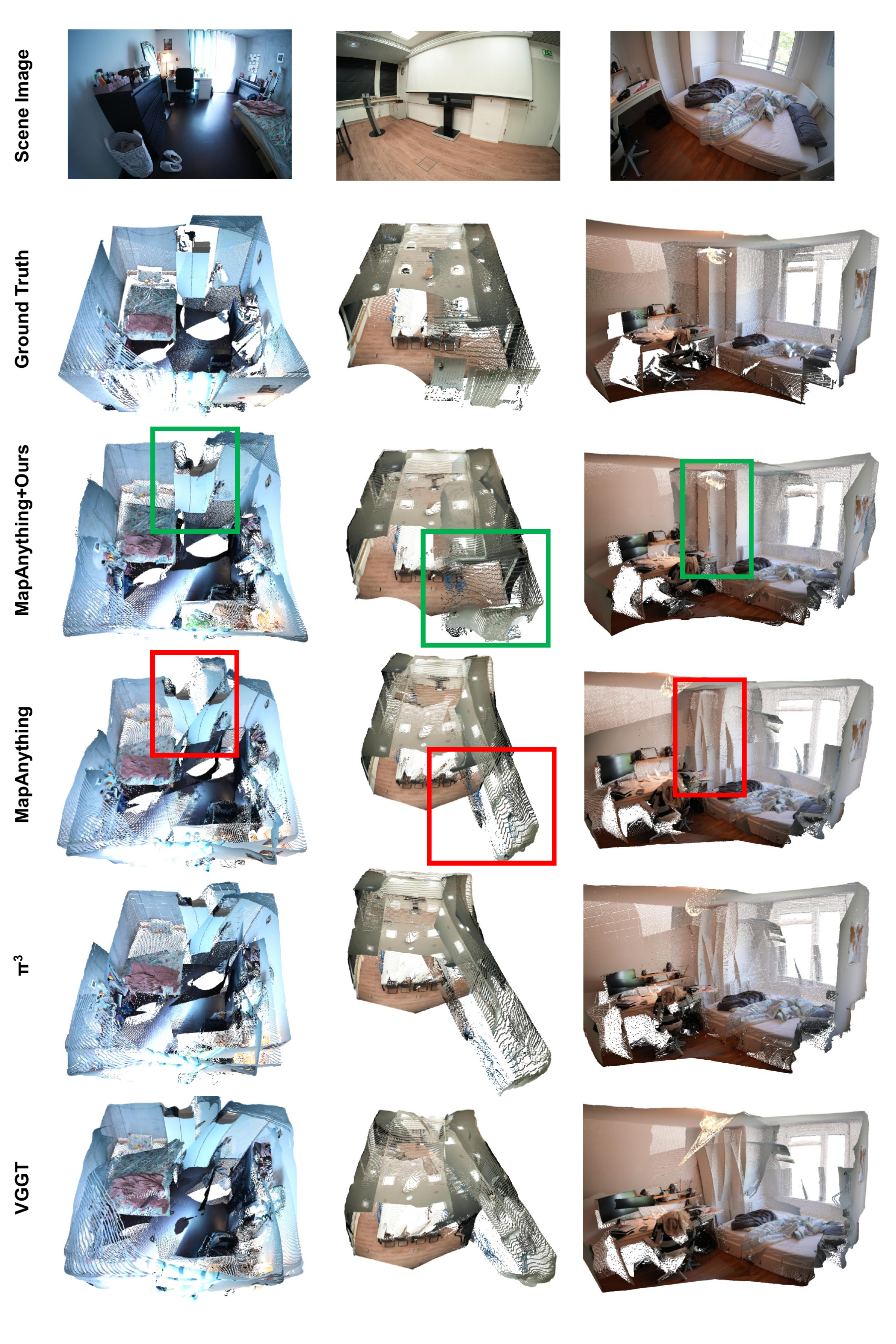}
    \caption{
        Qualitative examples of 3D reconstructions on ScanNet++.
    }
    \label{fig:vis_scannetpp}
\end{figure*}

\begin{figure*}[ht!]
    \centering
    \includegraphics[width=\linewidth]{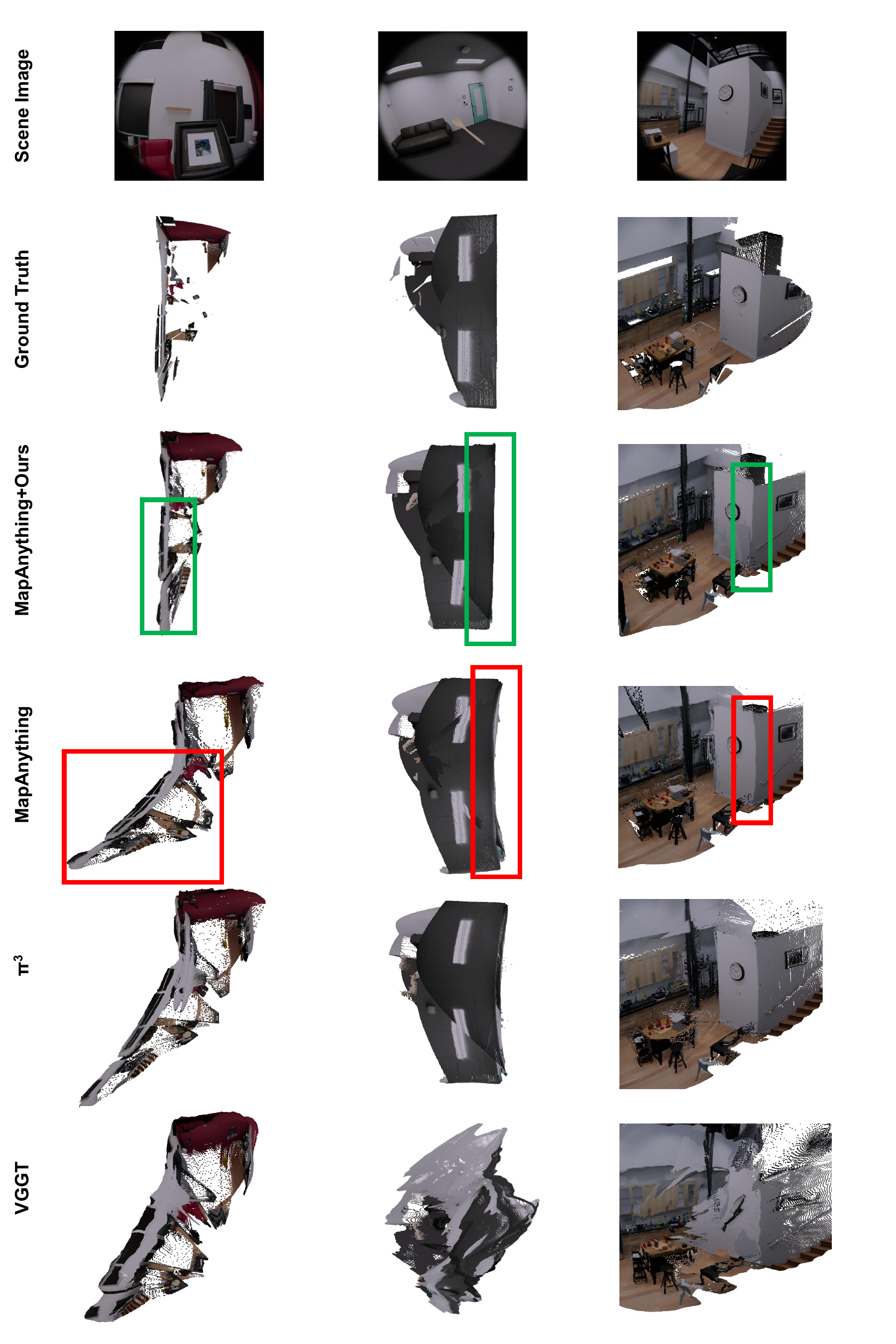}
    \caption{
        Qualitative examples of 3D reconstructions on ADT.
    }
    \label{fig:vis_adt}
\end{figure*}

\begin{figure*}[ht!]
    \centering
    \includegraphics[width=\linewidth]{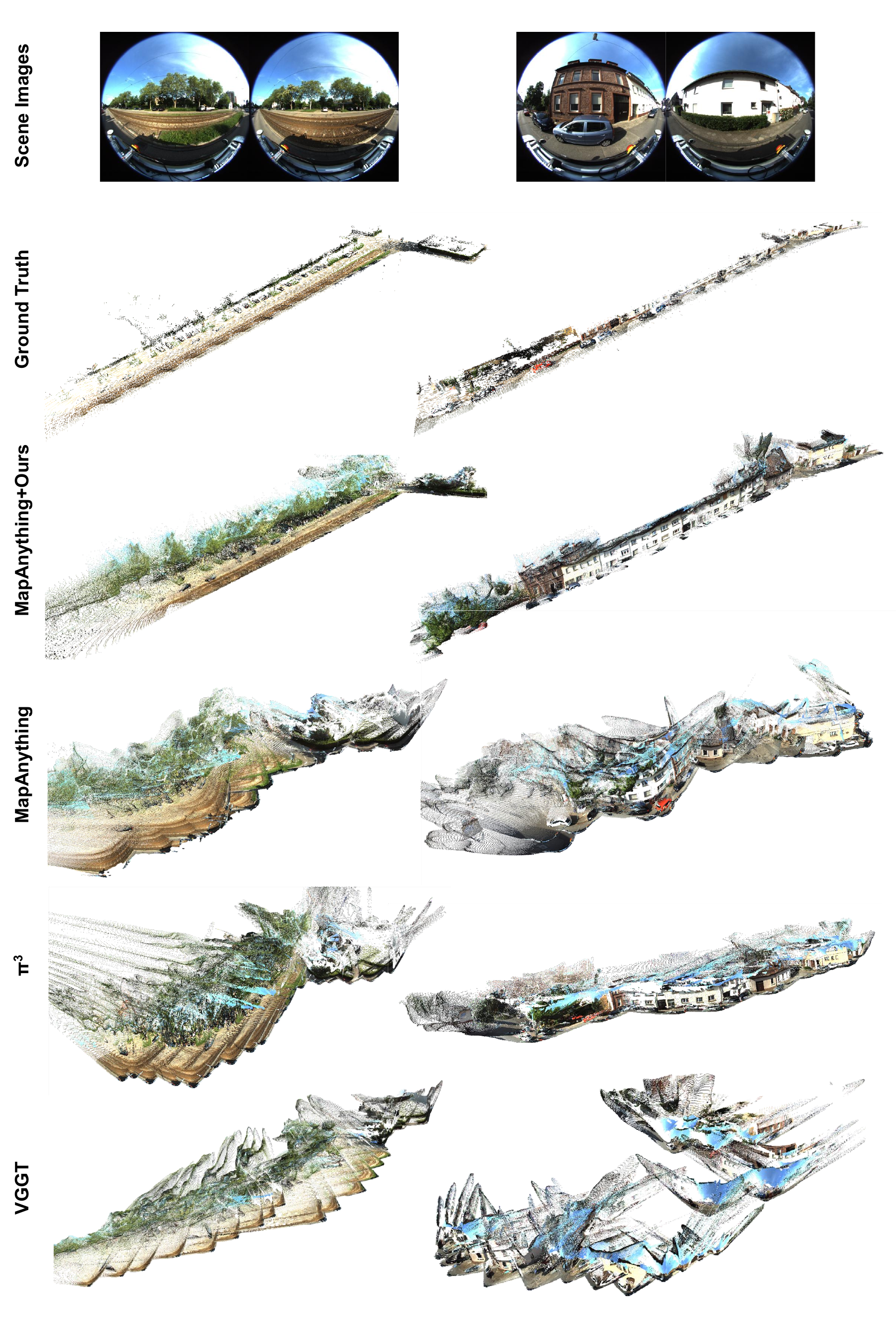}
    \caption{
        Qualitative examples of 3D reconstructions on KITTI360.
    }
    \label{fig:vis_kitti360}
\end{figure*}

\paragraphtitle{Additional qualitative results.}
We provide more visual examples of 3D reconstructions in \cref{fig:vis_scannetpp,fig:vis_adt,fig:vis_kitti360}.
In each figure, the first row presents some scene images, while the remaining ones show the ground-truth and predicted point clouds.
The green and red bounding boxes highlight regions where distortions are corrected by calibration tokens most significantly.
These examples demonstrate that Fisheye3R consistently improves geometric fidelity of the reconstructions, particularly in areas affected by strong lens distortion or challenging viewing angles.
Camera poses are also estimated more accurately, as reflected by the improved alignment of point maps from different views, especially on sequences with wider FoV and longer length (\cref{fig:vis_kitti360}).

\begin{table}[t]
    \centering
    \small
    \newcolumntype{M}{>{\centering\arraybackslash}m{15mm}}
    \begin{tabular}{c | MM}
    \toprule
        Method & Speed (FPS) & Memory (GiB) \\
        \midrule
        Baseline & 31.1 & 14.72 \\
        + Calibration Tokens & 31.1 & 14.72 \\
        + Calibration Tokens + Masked Attention & 29.1 & 15.26 \\
    \bottomrule
    \end{tabular}
    \vspace{1mm}
    \caption{
        Average inference frames per second (FPS) and peak GPU memory across the test sequences with MapAnything model and an A6000 GPU.
        The inference overhead is mainly from masked attention (6\% in runtime and 4\% in memory), which is only introduced when the test sequences have unknown and potentially mixed camera types.
    }
    \label{tab:efficiency}
    \vspace{-5mm}
\end{table}

\paragraphtitle{Training efficiency.}
For training, the peak GPU memory usage is reported in \cref{tab:ablation_components}, and the total training time is approximately 20 hours per model with 4 GPUs.
Moreover, the model can be trained using fully fisheye sequences instead of mixed-camera types, without employing masked attention during training, and masked attention can be applied only at inference.
We observe that this strategy speeds up training and has a negligible impact on the final performance on hybrid inputs.

\paragraphtitle{Inference efficiency.}
We report the inference efficiency of the baseline model before and after introducing calibration tokens and masked attention in \cref{tab:efficiency}, measured in frames per second (FPS) and peak GPU memory usage.
Calibration tokens themselves introduce negligible computational overhead in both runtime and memory, indicating that adapting the model to fully fisheye sequences can be achieved with essentially no additional computational cost.
Only when the camera type is unknown and mixed sequences must be supported is masked attention required to selectively activate calibration tokens, costing a modest overhead of approximately 6\% in runtime and 4\% in memory.

\begin{figure*}[t]
    \centering
    \includegraphics[width=\linewidth]{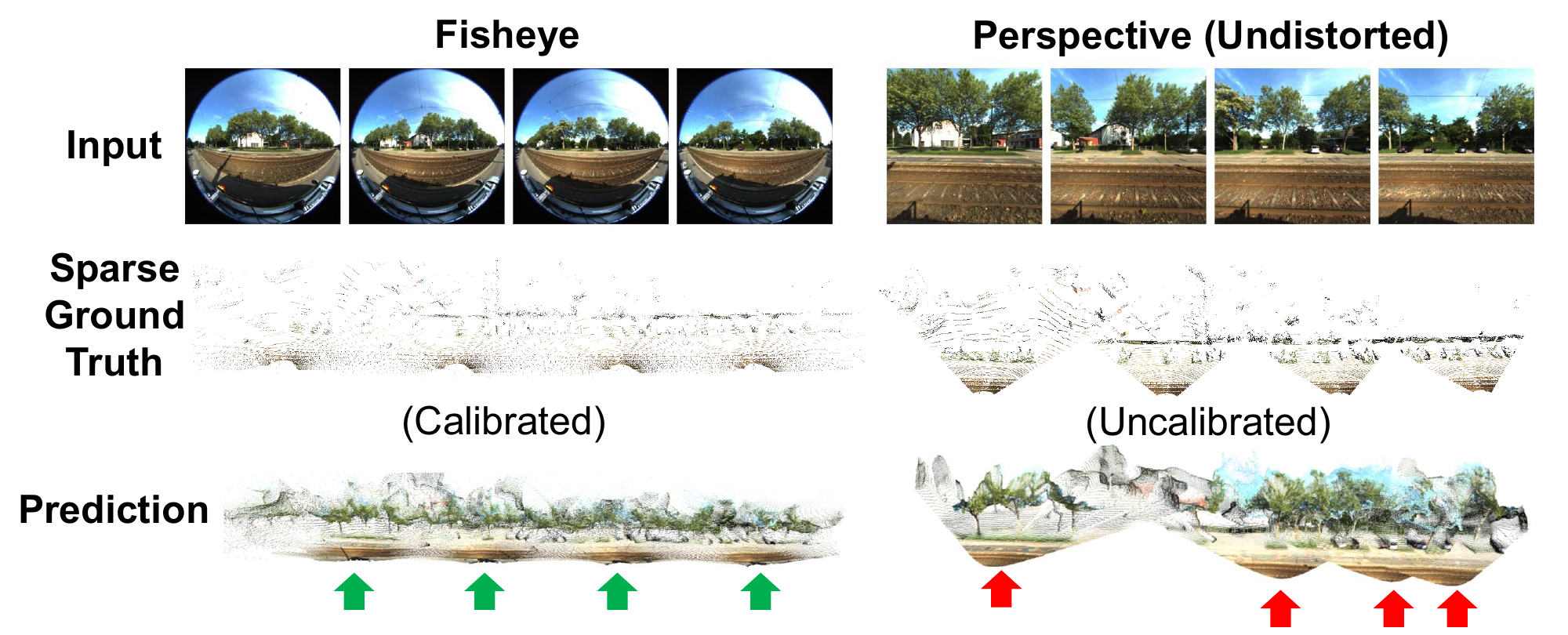}
    \vspace{-5mm}
    \caption{
        Qualitative comparison of 3D reconstruction on fisheye vs. undistorted imagery with MapAnything.
        Our calibrated model (left) directly processes raw fisheye sequences, yielding a geometrically consistent point cloud with accurate camera pose estimations (green arrows).
        In contrast, the uncalibrated baseline (right) operating on undistorted perspective images suffers from significant pose drift and geometric misalignment (red arrows) due to loss of cross-view covisibility from undistortion.
    }
    \label{fig:undistortion_supp}
\end{figure*}

\paragraphtitle{Ablation on calibrated modules.}
We study the effect of inserting calibration tokens into different components of the model in \cref{tab:ablation_components}.
All models are evaluated on fully fisheye sequences, where masked attention is not involved.
In this setting, one could in principle calibrate all $L_1=24$ layers of the image encoder to maximize the potential performance gain.
However, comparing the second (E$^*$) and third (E) rows of \cref{tab:ablation_components}, we observe no significant difference in performance, suggesting that calibrating the first $L_0 = 12$ layers provides little benefit, even when camera type classification is not required.
A possible explanation is that early-layer features have not yet captured sufficient distortion-related semantics for effective correction.

\paragraphtitle{Additional comparisons.}
We compare calibration tokens with LoRA \cite{hu2022lora} in \cref{tab:ablation_components}.
LoRA with different ranks is inserted to the same attention layers as calibration tokens, except the first $L_0 = 12$ encoder layers.
Overall, LoRA requires substantially more parameters for adaptation.
Particularly, LoRA with rank $r = 32$ introduces 24 times more training parameters than calibration tokens.
Both methods cost approximately 35GiB GPU memory.
For reconstruction, calibration tokens achieve the best pose and FoV estimation accuracy, while the two approaches perform similarly in depth and point map estimation.

\paragraphtitle{Fisheye undistortion.}
An alternative for processing fisheye imagery in 3D reconstruction is to perform undistortion to generate perspective images, which are then processed by the original uncalibrated foundation model.
However, this approach is fundamentally limited, as it requires precise prior calibration of camera parameters and introduces significant computational overhead.
More critically, the rectification process discards the wide FoV characteristic of fisheye lenses, often severing crucial cross-view covisibility that would otherwise exist in the native fisheye frames, as illustrated in \cref{fig:undistortion_supp}.
By shrinking the effective overlap between views, undistortion not only reduces total scene coverage but also destabilizes the geometric constraints required for pose estimation, leading to structural drift.
Therefore, developing a distortion-aware model is essential for efficient wide-area reconstruction.
A scene that traditionally requires hundreds of perspective images to ensure sufficient inter-frame overlap can be effectively modeled with tens of wide-FoV fisheye views (\cref{fig:vis_kitti360}).

\end{document}